\pdfoutput=1
\documentclass[11pt]{article}
\usepackage[final]{EMNLP2023}
\usepackage{times}
\usepackage{latexsym}
\usepackage{rotating}
\usepackage{graphicx}
\usepackage{pgfplots}
\usepackage{diagbox}
\usepackage{enumitem}
\usepackage{svg}
\usepackage{url}
\usepackage{collcell}
\usepackage{hyperref}
\usepackage{placeins}
\usepackage{pgf-pie}
\usepackage{tabularx}
\usepackage{tikz}
\usepackage{amssymb}
\usepackage{longtable}
\usepackage{mathrsfs}
\usepackage[font=small]{caption}
\usepackage[T1]{fontenc}
\usepackage[utf8]{inputenc}
\usepackage{microtype}
\usepackage{inconsolata}
\usepackage{booktabs}
\usepackage{multicol}
\usepackage{multirow}
\usepackage{arydshln}
\usepackage{makecell}
\usepackage{pdflscape}
\usepackage{fancybox}
\usepackage{pgfplotstable}
\pgfkeys{/pgf/number format/.cd,1000 sep={}}
\pgfplotsset{compat=1.17}
\usetikzlibrary{patterns}

\title{PMIndiaSum: Multilingual and Cross-lingual Headline Summarization \\for Languages in India}

\author{
Ashok Urlana\textsuperscript{1,*}\qquad\qquad
Pinzhen Chen\textsuperscript{2,*}\qquad\qquad
Zheng Zhao\textsuperscript{2}\\
\bf{}
Shay B. Cohen\textsuperscript{2}\qquad\quad
Manish Shrivastava\textsuperscript{1}\qquad\quad
Barry Haddow\textsuperscript{2}
\vspace{0.5ex}
\\
\vspace{0.5ex}
\textsuperscript{1}IIIT Hyderabad\qquad\qquad\textsuperscript{2}University of Edinburgh\\
\texttt{ashok.urlana@research.iiit.ac.in, \{pinzhen.chen,zheng.zhao\}@ed.ac.uk}\\
\texttt{scohen@inf.ed.ac.uk, m.shrivastava@iiit.ac.in, bhaddow@ed.ac.uk}
}

\newcommand{\mlen}{\mathrm{len}}

\begin{document}
\maketitle
\def\thefootnote{*}\footnotetext{Equal contribution.}\def\thefootnote{\arabic{footnote}}
\begin{abstract}
This paper introduces PMIndiaSum, a multilingual and massively parallel summarization corpus focused on languages in India. Our corpus provides a training and testing ground for four language families, 14 languages, and the largest to date with 196 language pairs. We detail our construction workflow including data acquisition, processing, and quality assurance. Furthermore, we publish benchmarks for monolingual, cross-lingual, and multilingual summarization by fine-tuning, prompting, as well as translate-and-summarize. Experimental results confirm the crucial role of our data in aiding summarization between Indian languages. Our dataset is publicly available and can be freely modified and re-distributed.\footnote{The corpus is released under the \href{https://creativecommons.org/licenses/by/4.0/}{CC BY 4.0 license} at \href{https://huggingface.co/datasets/PMIndiaData/PMIndiaSum}{https://huggingface.co/datasets/PMIndiaData/PMIndiaSum}\label{footnote:data-url}}
\end{abstract}

\section{Introduction}\label{sec:intro}

The era of deep learning has witnessed great advancements in various natural language processing (NLP) tasks. Yet prevalent solutions usually rely on large datasets, which are limited to high-resource languages. This is particularly pronounced in India, where languages spoken by a large population have been historically overlooked in research and under-resourced \cite{kumar-etal-2022-indicnlg}. 

In text summarization, where a system generates a brief description of a longer text, the availability of datasets for Indian languages is restricted in terms of both language coverage and size \citep{tacl-cross-lingual-survey}.\footnote{This paper refers to the languages widely used in India as ``languages in India'' or ``Indian languages'', but they are not exclusive to India and are spoken worldwide.} Moreover, some existing datasets are not easily accessible or have been criticized for their quality \cite{urlana-etal-2022-tesum}. Given the multilingual nature of India, having 122 major languages and 22 official ones, the development of cross-lingual summarization is desirable for information access, however, it is challenging due to the absence of reliable resources.

\begin{figure}[tb]
    \centering\footnotesize
    \includegraphics[width=0.4\textwidth,trim={0 0 0 5ex},clip]{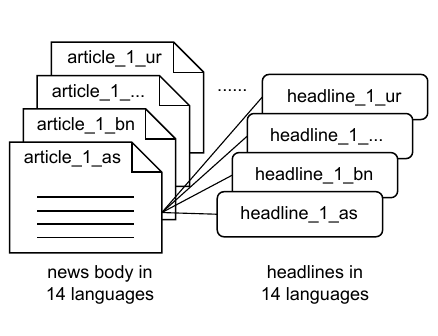}
    \caption{PMIndiaSum supports 14 languages and 196 language pairs, by sourcing from massively parallel articles.}
    \label{fig:a_nice_drawing}
\end{figure}

To address these challenges, we present PMIndiaSum, a massively multilingual and cross-lingual summarization dataset sourced from the Prime Minister of India website.\footnote{A governmental website that allows content re-distribution. \href{https://www.pmindia.gov.in/en/website-policies}{https://www.pmindia.gov.in/en/website-policies}} This website publishes political news, usually available in various languages covering the same content. In accordance with established methods \cite[][]{napoles-etal-2012-annotated2,rush-etal-2015-neural}, we use a news article as the document and its headline as the summary. Beyond monolingual data, as Figure~\ref{fig:a_nice_drawing} depicts, the website's multilingualism enables cross-lingual document-summary alignment with a firm confidence in quality.

Our efforts lead to an extensive coverage of 196 language directions from 14 languages across four families, making the corpus the current widest collection of Indian language pairs for summarization. There are 76,680 monolingual document-headline pairs and 620,336 cross-lingual pairs in total. We display language family and code information in Table~\ref{tab:language-families}. Of particular note is the inclusion of Manipuri (Meitei), an often neglected language in present-day datasets. The text in our corpus is in Bengali script. Besides the corpus release, we obtain results in multiple metrics for popular summarization paradigms like fine-tuning, two-step summarization-translation, and prompting, to serve as a reference for future research.

\begin{table}[t!]
\centering\small
\begin{tabularx}{0.47\textwidth}{lX}
\toprule
\multicolumn{1}{c}{\textbf{Family}} & \multicolumn{1}{c}{\textbf{Language (language code)}} \\
\midrule
Dravidian & Kannada (\texttt{kn}), Malayalam (\texttt{ml}), Tamil (\texttt{ta}), Telugu (\texttt{te}) \\
\midrule
Indo-Aryan & Assamese (\texttt{as}), Bengali (\texttt{bn}), Gujarati (\texttt{gu}), Hindi (\texttt{hi}), Marathi (\texttt{mr}), Odia (\texttt{or}), Punjabi (\texttt{pa}), Urdu (\texttt{ur})  \\
\midrule
Indo-European & English (\texttt{en}) \\
\midrule
Tibeto-Burman & Manipuri (\texttt{mni}) \\ 
\bottomrule
\end{tabularx}
\caption{Language family and language code information.}
\label{tab:language-families}
\end{table}

Though there exist summarization datasets supporting languages in India, some concerns need to be addressed with regard to developing datasets for Indian language summarization. Explicitly, data construction should (1) respect copyright permissions and ensure transparency in data processing; (2) carry out multifaceted quality checks; (3) offer a broad range of language directions; (4) provide reasonably sized data to facilitate system building. Our PMIndiaSum takes into consideration all of these matters. Following \citet{datasheet}'s advice, we include a datasheet in Appendix~\ref{appendix:data_sheet}.

\begin{table*}[ht]
\centering\small
\setlength{\tabcolsep}{1.1ex}
\begin{tabular}{lrrrrrrrrrrrrrr}
\toprule
& \multicolumn{1}{c}{\textbf{\texttt{as}}}
& \multicolumn{1}{c}{\textbf{\texttt{bn}}}
& \multicolumn{1}{c}{\textbf{\texttt{gu}}}
& \multicolumn{1}{c}{\textbf{\texttt{hi}}}
& \multicolumn{1}{c}{\textbf{\texttt{kn}}}
& \multicolumn{1}{c}{\textbf{\texttt{ml}}}
& \multicolumn{1}{c}{\textbf{\texttt{mni}}}
& \multicolumn{1}{c}{\textbf{\texttt{mr}}}
& \multicolumn{1}{c}{\textbf{\texttt{or}}}
& \multicolumn{1}{c}{\textbf{\texttt{pa}}}
& \multicolumn{1}{c}{\textbf{\texttt{ta}}}
& \multicolumn{1}{c}{\textbf{\texttt{te}}}
& \multicolumn{1}{c}{\textbf{\texttt{ur}}}
& \multicolumn{1}{c}{\textbf{\texttt{en}}} \\
\midrule
monolingual size & 2089 & 5557 & 6802 & 7363 & 5307 & 4665 & 4936 & 5816 & 4651 & 4814 & 6079 & 6126 & 4315 & 8160 \\
\hdashline
doc vocab & 73k & 130k & 125k & 94k & 166k & 211k & 143k & 147k & 98k & 80k & 150k & 202k & 61k & 68k \\
sum vocab & 6k & 10k & 14k & 10k & 12k & 12k & 12k & 12k & 9k & 9k & 13k & 16k & 8k & 11k \\
token/doc & 522 & 517 & 505 & 719 & 405 & 336 & 489 & 510 & 544 & 688 & 344 & 430 & 754 & 498 \\
sent/doc & 24.7 & 31.2 & 26.0 & 31.0 & 27.0 & 26.0 & 27.1 & 30.2 & 31.4 & 28.9 & 23.8 & 26.7 & 34.0 & 22.2 \\
token/sum & 11.9 & 11.5 & 12.3 & 15.4 & 12.3 & 10.1 & 13.5 & 11.4 & 10.8 & 18 & 11.7 & 14.1 & 17.7 & 13.4 \\
sent/sum & 1.0 & 1.0 & 1.0 & 1.0 & 1.0 & 1.0 & 1.0 & 1.0 & 1.0 & 1.0 & 1.0 & 1.0 & 1.0 & 1.0 \\
\hdashline
compression & 91.5 & 92.1 & 89.7 & 91.3 & 91.0 & 90.8 & 90.4 & 90.9 & 91.7 & 90.1 & 90.5 & 90.5 & 91.6 & 91.6 \\
density & 6.06 & 3.91 & 5.45 & 8.67 & 4.86 & 3.98 & 7.97 & 4.50 & 3.74 & 8.12 & 5.35 & 4.75 & 8.13 & 7.31 \\
novelty, 1gram & 27.9 & 30.0 & 27.4 & 12.7 & 28.5 & 30.7 & 22.4 & 29.5 & 29.2 & 15.2 & 24.7 & 26.9 & 13.1 & 22.9 \\
novelty, 2gram & 52.5 & 59.8 & 53.5 & 38.7 & 58.4 & 59.2 & 44.9 & 59.0 & 60.8 & 41.3 & 52.9 & 55.8 & 39.7 & 44.9 \\
novelty, 3gram & 63.5 & 73.3 & 65.9 & 54.1 & 70.9 & 72.2 & 59.8 & 71.1 & 73.5 & 57.6 & 66.1 & 71.9 & 56.3 & 56.9 \\
novelty, 4gram & 71.4 & 81.3 & 73.6 & 62.7 & 78.5 & 79.9 & 67.5 & 78.8 & 81.4 & 68.4 & 74.0 & 79.7 & 66.7 & 64.9 \\
redundancy, 1gram & 1.4  & 1.3 & 1.5 & 5.5 & 2.8 & 2.5 & 5.1 & 1.5 & 1.3 & 4.9 & 2.1 & 2.6 & 5.7 & 4.0 \\
redundancy, 2gram & 0.1 & 0.1 & 0.2 & 0.3 & 0.4 & 0.4 & 0.8 & 0.2 & 0.1 & 0.9 & 0.3 & 0.3 & 0.4 & 0.4 \\
\bottomrule
\end{tabular}
\caption{PMIndiaSum monolingual document-summary data analysis.}\label{tab:data_stats}
\end{table*}

\section{Preparation of PMIndiaSum}
\subsection{Data acquisition}
\label{sec:data_acquistion}
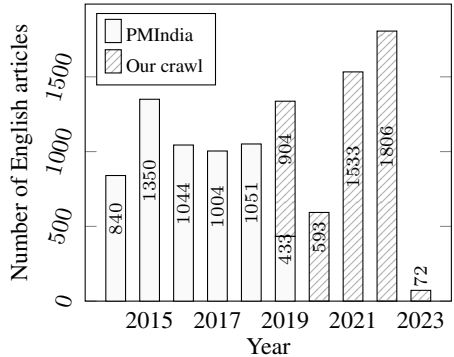
\begin{figure}[htb]
    \centering
    \pgfplotsset{every tick label/.append style={font=\footnotesize}}
\pgfplotsset{xlabel/.append style={font=\footnotesize}}
\pgfplotsset{ylabel/.append style={font=\footnotesize}}

\pgfplotstableread{
        year  pmindia ours
        2014  840    0
        2015  1350   0
        2016  1044   0
        2017  1004   0
        2018  1051   0
        2019  433    904
        2020  0      593
        2021  0      1533
        2022  0      1806
        2023  0      72
}\datalabel

\begin{tikzpicture}
    \begin{axis}[
        width=0.4\textwidth,
        ybar stacked,
        ylabel style={inner sep=0ex},
        xtick=data,
        xtick style={draw=none},
        bar width=1.5ex,
        legend style={cells={anchor=west}, nodes={scale=0.8}, legend pos=north west},
        xticklabels={,2015,,2017,,2019,,2021,,2023},
        yticklabel style={rotate=75},
        xlabel={Year},
        ymin=0,
        xlabel style={inner sep=0ex},
        ylabel={Number of English articles},
        nodes near coords,
        nodes near coords style={font=\scriptsize,rotate=90,xshift=1.5ex}
    ]
    \addplot [fill=black!2] table [y=pmindia, x expr=\coordindex] {\datalabel};
    \addlegendentry{PMIndia}
    \addplot [pattern=north east lines,pattern color=black!30] table [y=ours, x expr=\coordindex] {\datalabel};
    \addlegendentry{Our crawl}
    \end{axis}
\end{tikzpicture}
    \caption{PMIndiaSum acquisition statistics for English, where 56\% are from PMIndia and 44\% are newly crawled.}
    \label{fig:crawling_counts}
\end{figure}

The Prime Minister of India website posts articles that consist of a headline and a news body. These articles are available in multiple languages, with English being the default option. The HTML structure of each article includes a language indicator and a pointer to the English version. We gather website data for all articles in all available languages, which are sourced from two channels:
\setlist{nolistsep}
\begin{enumerate}\setlength\itemsep{0ex}\setlength{\itemindent}{-0.5ex}
    \item We ingest readily crawled HTML data from the PMIndia parallel corpus release \cite{haddow2020pmindia}, which correspond to articles published between 2014 and 2019.
    \item We newly crawl for more articles up to early 2023, using PMIndia crawler.\footnote{\href{https://github.com/bhaddow/pmindia-crawler}{https://github.com/bhaddow/pmindia-crawler}}
\end{enumerate}
\noindent We design a parser to retrieve article bodies and headlines from the HTML. We harvest 94,036 raw article-headline pairs for all languages, which we preliminarily regard as monolingual document-summary data. Figure~\ref{fig:crawling_counts} outlines the origins of English articles across different years.

\subsection{Data processing}
\label{sec:data_processing}

To create a high-quality dataset, the collected headline-body pairs undergo rule-based processing in the context of monolingual summarization. The cleaned version contains 76,680 monolingual instances for all languages, which is 81.5\% of the raw size. We describe the filtering rules, and list them in Appendix~\ref{sec:filtering_numbers} Table~\ref{tab:filtration_counts} detailing the number of invalid samples removed at each step.

\paragraph{Language mismatch} Despite confidence in the website's efforts towards language correctness, we set our own language filtering. We discard an entire data instance if either the document or the headline contains text outside of the Unicode range of its designated language.\footnote{Defined under the South and Central Asia-I languages on \href{https://unicode.org/versions/Unicode13.0.0/}{https://unicode.org/versions/Unicode13.0.0/}} We notice that a large number of samples removed, especially from \texttt{bn} and \texttt{en}, are code-mixed data.

\paragraph{Duplication and empty} To maintain a dataset with unique document-summary pairs, we remove all duplicates. We also eliminate samples that have identical summaries to steer clear of any headline-article mismatch errors from the website. In addition, we take out instances if either their document or summary is empty.

\paragraph{Prefix} To enforce that PMIndiaSum is abstractive in nature, we remove all samples where the summary is repeated as the initial or first few sentences in the document. For cases where the summary appears later, the model still needs to “understand” the document in order to “select” a summary. Here, such extraction becomes an easier (but valid) case of summarization.

\paragraph{Length} We filter out data pairs where the document contains less than two sentences or the summary contains less than three tokens, as these pairs may not provide sufficient information for training a summarization model and can become outliers that negatively impact a system's performance. We used the \texttt{indic\_nlp\_library} for both word and sentence segmentation \citep{kunchukuttan2020indicnlp}.

\subsection{Monolingual statistics}
Table~\ref{tab:data_stats} presents the demographics of PMIndiaSum for qualitative inspection. We use token-based measures \citep{grusky-etal-2018-newsroom2,narayan-etal-2018-dont}. Since many are not well-defined for cross-lingual data, we report the figures for monolingual pairs.

\paragraph{Vocabulary} We list the count of unique tokens in documents and summaries for each language, under the same tokenization as earlier.

\paragraph{Length and compression.} Our average document length is 27 sentences with 518 tokens with \texttt{ur} being the longest, and \texttt{ml} being the shortest. Summaries are on average 12 tokens, with nearly all being a single sentence. We then compute the compression ratio, which quantifies how concise a summary $S$ is given its document $D$ as $1-\frac{\mlen(S)}{\mlen(D)}$ in \citep{bommasani-cardie-2020-intrinsic2}. For each language, we display the average over all samples. A high compression of around 90\% implies that the headlines are extremely abstractive and condensed.

\paragraph{Density} \citet{grusky-etal-2018-newsroom2} introduced extractive fragments $F(D,S)$ as a set of shared sequences in document $D$ and summary $S$. We use their density metric which reflects the degree to which the summary can be composed of texts from the document: $\mathit{density}=\frac{1}{\mlen(S)} {\sum_{\textit{f}\in{F(D,S)}}}\mlen(f)^2$.

\paragraph{Novelty} To evaluate the originality of summaries in our corpus, we compute the percentage of $n$-grams that are present in a summary but not in its corresponding document. We report 1-to-4-gram novelty scores averaged across all samples in each language. Here, unigram novelty is equivalent to $1-\mathit{coverage}$ in \citep{grusky-etal-2018-newsroom2}.

\paragraph{Redundancy} \citet{hasan-etal-2021-xl} described redundancy as the amount of repeated information in a summary. It is calculated as the ratio of repetitive n-grams to all $n$-grams in the summary. We measure the average redundancy for unigrams and bigrams, where lower is more desirable.

\subsection{Multilingualism and parallelism}
\begin{figure}[ht]
    \centering
    \pgfplotsset{every tick label/.append style={font=\scriptsize}}
\pgfplotsset{xlabel/.append style={font=\footnotesize}}
\pgfplotsset{ylabel/.append style={font=\footnotesize}}
\pgfplotstableread{
        lang    count
        1       611
        2       1091
        3       473
        4       670
        5       637
        6       729
        7       977
        8       1078
        9       984
        10      986
        11      980
        12      551
        13      570
        14      232
}\datalabel

\begin{tikzpicture}
    \begin{axis}[
        ybar,
        width=0.5\textwidth,
        height=0.3\textwidth,
        ylabel style={inner sep=0ex},
        xtick=data,
        nodes near coords,
        yticklabel style={rotate=75},
        xlabel={Number of languages available},
        xticklabels={1,2,3,4,5,6,7,8,9,10,11,12,13,14},
        ymin=0,
        ymax=1200,
        xtick style={draw=none},
        bar width=1.5ex,
        xlabel style={inner sep=0ex},
        xtick align=inside,
        ylabel={Number of articles},
        nodes near coords style={font=\scriptsize,rotate=90, anchor=east}
    ]
    \addplot[] table [y=count, x expr=\coordindex] {\datalabel};
    \end{axis}
\end{tikzpicture}
    \caption{Degree of article parallelism in PMIndiaSum.}\label{fig:parallel_counts}
\end{figure}
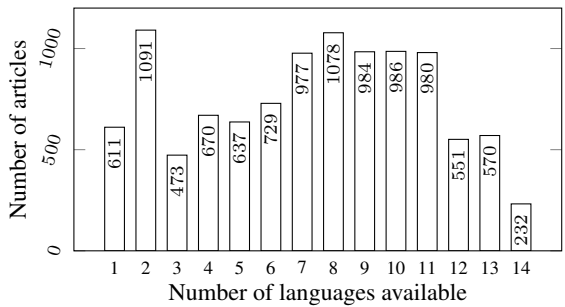

On top of monolingual data, PMIndiaSum features massive parallelism because the source website publishes a single article in multiple language versions to cater for its audience. This means that document-headline pairs in various languages derived from the same article enable cross-lingual summarization. As depicted in Figure~\ref{fig:parallel_counts}, most of the articles are available in at least two languages, and 232 articles are available in all languages. This allows for the creation of cross-lingual and multilingual summarization data pairs.

Technically, every document is paired with summaries in other languages from the same article, and vice versa. Matching is done via the default English pointer in HTML. Such multi-way parallelism results in $14\times{}13=182$ cross-lingual pairs in addition to monolingual data. The average data size is 5,477 for monolingual and 3,408 for cross-lingual summarization. Appendix~\ref{appendix:data_counts} Table~\ref{tab:cls_counts} details the sizes for all 196 language directions.

\subsection{Data split}
\label{sec:data_split}
To prevent test data leakage in multilingual models, where a model sees an article in one language and is tested on the same article in a different language, we isolate the 232 articles that are available in all languages for validation and testing. We divide the data equally into validation and test sets, resulting in 116 instances for each language pair in each set. This approach provides consistent validation and test splits for each target summary language, regardless of the language directions. All other data are in the training split.

\subsection{Quality considerations}
We have high confidence in text quality since the corpus is crawled from a governmental website and we have specifically designed an HTML parser to remove extraneous elements. In addition, we discuss candidate summaries and data parallelism.

\subsubsection{Summary choices}\label{sec:summary_quality}

We take \citet{rush-etal-2015-neural}'s approach of treating the headline as an article's summary, whereas some works opt for the first sentence instead, e.g. XSum and XL-Sum \citep{narayan-etal-2018-dont,hasan-etal-2021-xl}. Although the lead sentence can be an overview of an article, this paradigm has received criticism due to potential concerns \citep{zhao-etal-2020-reducing,urlana-etal-2022-tesum}:
\setlist{nolistsep}
\begin{enumerate}\setlength\itemsep{0ex}\setlength{\itemindent}{-0.5ex}
    \item The first sentence can be part of multi-sentence writing, which is difficult to isolate.
    \item Further, the second sentence onwards, when employed as a document, may not always contain all the information in the first sentence.
\end{enumerate}

We conduct an empirical study on the suitability of headlines versus first sentences as summaries. For each of English, Hindi, and Telugu, we invite three native speakers to perform a manual inspection on 50 random samples. Given an article, we ask if the first sentence is a summary and if the headline is a summary. We record majority votes for each item in Table~\ref{tab:human_eval_yes_no}, which suggests that headlines are consistently regarded as a summary; on the contrary, first sentences may not always function as a summary, which is also hard to be automatically identified. In Appendix~\ref{appendix:data_example}, we outline the evaluation protocol and supply a few examples of problematic first sentences. Based on the evaluation outcome, we determine that it is more appropriate to use the headlines as summaries in our corpus. We argue that having the high parallelism between articles, findings from three languages can generalize to the whole corpus.

\subsubsection{Parallelism}
Finally, we assert the validity of cross-lingual alignments. We measure the degree of parallelism by calculating cosine similarity between neural presentations of texts in two languages \citep[LaBSE,][]{feng-etal-2022-language2}. We compute LaBSE scores between summaries as well as between entire documents, from the same article, for each language pair. The average cross-lingual LaBSE score is 0.86 for summaries and 0.88 for documents. These scores indicate the high parallelism between summaries and between documents in different languages; they also notably exceed the 0.74 summary-summary threshold \citet{Bhattacharjee2022crosssum} used to extract CrossSum from XL-Sum. Hence, given monolingual document-summary pairs, our choice of substituting the summary (or the document) with one in another language maintains the integrity of cross-lingual pairs. We enclose all pairwise LaBSE scores in Appendix~\ref{appendix:labse_scores} Table~\ref{tab:labse_scores} for reference.

\begin{table}[tb]
    \centering\small
    \begin{tabular}{lccc}
\toprule
& \texttt{\textbf{en}} & \texttt{\textbf{hi}} & \texttt{\textbf{te}} \\
\midrule
    Headlines       & \textbf{47} & \textbf{49} & \textbf{49} \\
    First sentences & 28          & 45          & 35          \\
\bottomrule
    \end{tabular}
    \caption{Number of times, out of 50, headlines and first sentences are considered a summary by evaluators.}\label{tab:human_eval_yes_no}
\end{table}

\section{Benchmark Experiments}
\label{sec:benchamrk_experiments}
\subsection{Task and evaluation}

Formally, given a source document $D$, the process of summarization should produce a target summary $S$ with a shorter length, yet conveying the most important message in $D$. We explore three types of models defined by language directions:
\begin{enumerate}\setlength\itemsep{0ex}\setlength{\itemindent}{-0.5ex}
    \item Monolingual: document $D_{L}$ and summary $S_{L}$ are in the same language $L$.
    \item Cross-lingual: document $D_{L}$ and summary $S_{L'}$ are in different languages $L$ and $L'$.
    \item Multilingual: monolingual and cross-lingual summarization from $D_{\{L_1,L_2,...,L_n\}}$ to $S_{\{L_1,L_2,...,L_n\}}$ within a single model.
\end{enumerate}

\noindent  In our context, cross-lingual models summarize from one single language to another. Monolingual and cross-lingual models could be more accurate as they concentrate on one language (pair), whereas multilingual models can significantly save storage and computational resources, and likely transfer knowledge across languages.

For evaluation, we report F1 scores of ROUGE-2/L \cite{lin-2004-rouge}, as implemented in XL-Sum, with default language-specific settings such as segmentation and stemming.\footnote{\href{https://github.com/csebuetnlp/xl-sum/tree/master/multilingual\_rouge\_scoring}{https://github.com/csebuetnlp/xl-sum/tree/master/multilingual\_rouge\_scoring}} We also use BLEU-4 from \texttt{sacrebleu} \citep{papineni-etal-2002-bleu,post-2018-call}.\footnote{nrefs:1|case:mixed|eff:no|tok:13a|smooth:exp|version:2.3.1} We abbreviate these as R2, RL, and BL hereafter.

\subsection{Methodology}
We intend to provide a benchmark for conventional summarization approaches to cover all language scenarios. We introduce the paradigms below and list the language settings tested with each method in Table~\ref{tab:methodology}.

\begin{table}[th]
\centering\small
\begin{tabular}{lccc}
\toprule
& \textbf{Mono} & \textbf{Cross} & \textbf{Multi} \\
\midrule
Extractive: lead  & \checkmark & & \\
Extractive: oracle  & \checkmark & & \\
Summarize-then-translate & & \checkmark & \\
Translate-then-summarize & & \checkmark & \\
Fine-tuning: full & \checkmark & \checkmark & \checkmark \\
Fine-tuning: zero-shot &  & \checkmark & \\
LLM prompting & \checkmark & & \\
\bottomrule
\end{tabular}
\caption{Summarization approaches and language directions.}
\label{tab:methodology}
\end{table}

\begin{table*}[htb]
\centering\small
\setlength{\tabcolsep}{0.825ex}
\begin{tabular}{rcccccccccccc} 
\toprule
& \multicolumn{3}{c}{\textbf{Lead}} & \multicolumn{3}{c}{\textbf{Oracle}} & \multicolumn{3}{c}{\textbf{IndicBART}} & \multicolumn{3}{c}{\textbf{mBART}} \\
\cmidrule(lr){2-4}\cmidrule(lr){5-7}\cmidrule(lr){8-10}\cmidrule(lr){11-13}
 & R2 & RL & BL & R2 & RL & BL & R2 ($\Delta$) & RL ($\Delta$) & BL ($\Delta$) & R2 ($\Delta$) & RL ($\Delta$) & BL ($\Delta$) \\
\midrule
\textbf{\texttt{as}} & 31.7 & 41.6 & 18.2 & 34.5 & 44.7 & 23.1 & \textbf{41.7} (15.4$\uparrow$) & \textbf{56.1} (18.5$\uparrow$) & \textbf{35.2} (17.4$\uparrow$) & - & - & - \\
\textbf{\texttt{bn}}& 27.4 & 40.3 & 13.5 & \textbf{31.0} & 43.7 & 18.8 & 27.8 (10.5$\uparrow$) & 46.2 (16.9$\uparrow$) & 20.5 (10.1$\uparrow$) & 30.5 (1.3$\uparrow$) & \textbf{48.2} (0.3$\uparrow$) & \textbf{21.5} (1.1$\uparrow$)  \\
\textbf{\texttt{gu}} & 33.7 & 45.3 & 22.7 & 38.4 & 50.9 & 24.6 & 49.8 \phantom{0}(4.4$\uparrow$) & 64.6 \phantom{0}(6.7$\uparrow$) & 44.6 \phantom{0}(8.0$\uparrow$)& \textbf{50.1} (1.8$\uparrow$) & \textbf{64.8} (0.3$\downarrow$) & \textbf{45.2} (4.0$\uparrow$) \\
\textbf{\texttt{hi}} & 44.8 & 58.2 & 30.1 & 49.1 & 62.5 & 33.2 & 53.1 \phantom{0}(4.6$\uparrow$) & 66.9 \phantom{0}(6.3$\uparrow$) & 46.0 \phantom{0}(7.3$\uparrow$) & \textbf{55.9} (1.4$\uparrow$) & \textbf{69.4} (0.4$\uparrow$) & \textbf{48.6} (1.3$\uparrow$)\\
\textbf{\texttt{kn}} & 26.4 & 37.4 & 18.1 & 31.1 & 42.8 & 19.7 & \textbf{39.9} \phantom{0}(4.3$\uparrow$) & \textbf{56.2} \phantom{0}(7.8$\uparrow$) & \textbf{36.3} \phantom{0}(7.8$\uparrow$) & - & - & - \\
\textbf{\texttt{ml}} & 23.8 & 37.1 & 12.7 & \textbf{32.7} & 45.8 & \textbf{15.4} & 30.3 \phantom{0}(6.3$\uparrow$) & \textbf{47.5} \phantom{0}(7.5$\uparrow$) & 15.3 \phantom{0}(5.1$\uparrow$) & 30.2 (3.1$\uparrow$) & 47.4 (1.8$\uparrow$) & 14.6 (1.2$\downarrow$) \\
\textbf{\texttt{mni}} & 38.3 & 50.5 & 26.4 & \textbf{42.0} & 54.2 & 26.0 & 38.7 \phantom{0}(0.7$\downarrow$) & 53.0 \phantom{0}(1.2$\downarrow$) & 32.0 \phantom{0}(0.8$\downarrow$) & 41.3 (1.5$\uparrow$) & \textbf{56.4} (0.9$\uparrow$) & \textbf{35.0} (1.6$\uparrow$)\\
\textbf{\texttt{mr}}& 31.0 & 45.2 & 20.8 & 33.5 & 48.7 & 21.8 & 42.9 (10.5$\uparrow$) & 59.6 (14.4$\uparrow$) & 39.7 (14.7$\uparrow$) & \textbf{43.2} (3.0$\uparrow$) & \textbf{61.1} (3.1$\uparrow$) & \textbf{39.8} (4.0$\uparrow$) \\
\textbf{\texttt{or}} & 21.8 & 37.3 & 11.2 & 26.2 & 42.0 & 15.3 & \textbf{38.6} \phantom{0}(3.0$\uparrow$) & \textbf{55.4} \phantom{0}(6.2$\uparrow$) & \textbf{28.7} \phantom{0}(5.1$\uparrow$) & - & - & - \\
\textbf{\texttt{pa}} & 44.0 & 57.7 & 30.7 & 47.1 & 60.7 & 30.3 & \textbf{54.2} \phantom{0}(3.4$\uparrow$) & \textbf{68.5} \phantom{0}(4.3$\uparrow$) & \textbf{47.5} \phantom{0}(4.5$\uparrow$) & - & - & -\\
\textbf{\texttt{ta}} & 27.9 & 41.6 & 22.9 & 38.0 & 52.8 & 25.5 & \textbf{47.5} \phantom{0}(6.5$\uparrow$)& 62.2 \phantom{0}(5.8$\uparrow$) & 37.4 \phantom{0}(5.8$\uparrow$)& 47.4 (5.8$\uparrow$) & \textbf{63.3} (4.1$\uparrow$) & \textbf{37.8} (2.7$\uparrow$)\\
\textbf{\texttt{te}} & 31.2 & 41.0 & 18.0 & \textbf{34.4} & \textbf{45.2} & \textbf{19.5} & 16.3 \phantom{0}(3.0$\uparrow$) & 32.7 \phantom{0}(3.0$\uparrow$)  & \phantom{0}9.9 \phantom{0}(1.9$\uparrow$) & 16.0 (0.4$\uparrow$) & 33.4 (0.0)\phantom{$\uparrow$} & \phantom{0}9.8 (0.8$\uparrow$)\\
\textbf{\texttt{ur}} & 21.6 & 30.2 & 18.1 & 39.9 & 52.7 & 24.9 & - & - & - & \textbf{52.6} (4.3$\uparrow$) & \textbf{64.8} (2.6$\uparrow$) & \textbf{44.0} (3.1$\uparrow$) \\
\textbf{\texttt{en}} & 33.3 & 42.1 & 18.6 & 38.7 & 48.2 & 22.8 & 63.4 \phantom{0}(5.5$\uparrow$) & 74.5 \phantom{0}(5.3$\uparrow$) & 53.1 \phantom{0}(7.7$\uparrow$) & \textbf{66.9} (2.9$\uparrow$) & \textbf{77.8} (1.0$\uparrow$) & \textbf{57.2} (5.6$\uparrow$)\\
\bottomrule
\end{tabular}
\caption{Monolingual benchmarks: separate models for each language. Bold indicates the best result; ($\Delta$) refers to the change relative to the corresponding multilingual result in Table~\ref{tab:multilingual-indicbart} or Table~\ref{tab:multilingual-mbart}.}\label{tab:mono_results2}
\end{table*}

\paragraph{Extractive baseline} We employ two training-free baselines: 1) selecting the lead sentence, and 2) scoring each sentence in the document against the reference and picking the best in an oracle way.

\paragraph{Fine-tuning} Fine-tuning pre-trained language models (PLMs) has shown promising results in summarization for Indian languages \citep{taunk2022summarizing, urlana2022indian}. The paradigm is to load a PLM and further train it for summarization with our data. In monolingual and cross-lingual settings, a PLM is only fine-tuned for a single language direction. On the other hand, in the multilingual setting, we simply mix and shuffle data for all language pairs, because the data sizes are of the same magnitude across all directions.

\paragraph{Summarization-and-translation} In cross-lingual settings, it is practical to leverage a translation system for language conversion. Two common pipelines are 1) summarize-then-translate, where a document is summarized in its original language and then the summary is translated into the target language, and 2) translate-then-summarize, which means first translating the document into the target language followed by summarizing it in that language.

\paragraph{Zero-shot fine-tuning} We fine-tune a PLM on all monolingual data only. This serves as an ablation study which only allows same-language document-headline data, but not cross-lingual. We then perform cross-lingual summarization on the test set with this model to measure its zero-shot capability.

\paragraph{Prompting}
Large language models (LLMs) have demonstrated potential in automatic summarization via prompting \cite{zhang2023benchmarking}. On our data, we give a preliminary evaluation of two instruction-finetuned LLMs based on LLaMA-7B \cite{touvron2023llama}: Alpaca \cite{alpaca2023} and Vicuna \cite{vicuna2023}. We query an LLM with a prompt ``\texttt{Article: \$\{article\}. Summarize the article above into a headline in \$\{language\} language. Summary:}'', and parse the model's completion as the candidate summary.
\begin{table}[t!]
\centering
\small
\begin{tabular}{clccccc} 
\toprule
 & & R2 & RL & BL   \\
\midrule
\multirow{2}{*}{fine-tuning} & {IndicBART} & 63.4 & 74.5 & 53.1  \\
& {mBART} & 66.9 & 77.8 & 57.2  \\
\hdashline
\multirow{2}{*}{prompting} & {Alpaca-7B} & 20.5 & 35.8  & 14.6  \\
& {Vicuna-7B} & 30.7 & 48.1 & 16.4  \\
\bottomrule
\end{tabular}
\caption{LLM prompting results for English.}
\label{tab:llm_results}
\end{table}

\paragraph{Other techniques} Lastly, we provide a preliminary discussion on using adapters and summarization models trained on other datasets in Appendix~\ref{appendix:adapter_other_model}.

\subsection{Systems}
Our fine-tuning paradigm is tested on two PLMs: IndicBART \cite{dabre-etal-2022-indicbart} and mBART-50 \cite{tang-etal-2021-multilingual}.\footnote{\href{https://huggingface.co/ai4bharat/IndicBARTSS}{https://huggingface.co/ai4bharat/IndicBARTSS}. We use the IndicBARTSS variant, which deals each language in its own script without the need of mapping to or from Devanagari.}\textsuperscript{,}\footnote{\href{https://huggingface.co/facebook/mbart-large-50}{https://huggingface.co/facebook/mbart-large-50}} We follow each PLM's convention to add language identification tokens to inform the PLM of the source and target languages. The PLMs support various languages, yet neither supports Manipuri; therefore, we randomly initialize an embedding entry as the \texttt{mni} language token in both IndicBART and mBART.\footnote{\texttt{mni} and \texttt{ur} are not covered by IndicBART; \texttt{as}, \texttt{kn}, \texttt{mni}, \texttt{or}, and \texttt{pa} are not supported by mBART.}

In the summarization-and-translation workflow, monolingual summarization is done using our monolingual fine-tuned PLMs introduced above. The translation process is delegated to a public engine that supports all the involved languages.\footnote{\href{https://ssmt.iiit.ac.in/translate}{https://ssmt.iiit.ac.in/translate}}

Training configurations are detailed in Appendix~\ref{appendix:exp_setup}. All trainable experiments are conducted three times to obtain the mean and standard deviation statistics. We report average scores, and attach standard deviations in Appendix~\ref{appendix:std_scores}.

\begin{table*}[htb]
\centering\small
\setlength{\tabcolsep}{0.85ex}
\begin{tabular}{lcccccccccccccccccccc}
\toprule
 & \multicolumn{4}{c}{\textbf{en-en}} &
  \multicolumn{4}{c}{\textbf{hi-hi}} &
  \multicolumn{4}{c}{\textbf{te-te}} &
  \multicolumn{4}{c}{\textbf{en-hi}} &
  \multicolumn{4}{c}{\textbf{hi-en}} \\
\cmidrule(lr){2-5}\cmidrule(lr){6-9}\cmidrule(lr){10-13}\cmidrule(lr){14-17}\cmidrule(lr){18-21}
 & \multicolumn{2}{c}{Mono} &
  \multicolumn{2}{c}{Multi} &
  \multicolumn{2}{c}{Mono} &
  \multicolumn{2}{c}{Multi} &
  \multicolumn{2}{c}{Mono} &
  \multicolumn{2}{c}{Multi} &
  \multicolumn{2}{c}{Cross} &
  \multicolumn{2}{c}{Multi} &
  \multicolumn{2}{c}{Cross} &
  \multicolumn{2}{c}{Multi} \\
\cmidrule(lr){2-3}\cmidrule(lr){2-3}\cmidrule(lr){4-5}\cmidrule(lr){6-7}\cmidrule(lr){8-9}\cmidrule(lr){10-11}\cmidrule(lr){12-13}\cmidrule(lr){14-15}\cmidrule(lr){16-17}\cmidrule(lr){18-19}\cmidrule(lr){20-21}
 & \multicolumn{2}{c}{IB \ mB} & \multicolumn{2}{c}{IB \ mB} & \multicolumn{2}{c}{IB \ mB} & \multicolumn{2}{c}{IB \ mB} & \multicolumn{2}{c}{IB \ mB} & \multicolumn{2}{c}{IB \ mB} & \multicolumn{2}{c}{IB \ mB} & \multicolumn{2}{c}{IB \ mB} & \multicolumn{2}{c}{IB \ mB} & \multicolumn{2}{c}{IB \ mB} \\
\midrule
{\textbf{Comprehensibility}} & 1  & 0  & {6}  & 0  & 0  & 0  & {9}  & 0  & 2  & {6}  & 5  & 5  & 1  & 1  & \textbf{39} & 0  & 4  & 2  & \textbf{31} & 1  \\
{\textbf{Grammar \& Fluency}}           & 0  & 0  & 1  & 0  & 2  & 1  & 0  & 1  & 1  & 1  & 0  & 0  & 0  & 1  & 0  & 1  & 1  & 2  & 1  & 1  \\
{\textbf{Factuality}}        & 2  & 1  & 1  & 1  & 4  & 4  & 3  & 3  & 6  & 6  & 6  & 5  & \textbf{22} & \textbf{26} & 3  & 7  & \textbf{21} & \textbf{18} & 6  & 10 \\
{\textbf{Omission}}          & 18 & 14 & 11 & 15 & \textbf{23} & \textbf{22} & 12 & 19 & 12 & 18 & 16 & 17 & 11 & 12 & 1  & 15 & \textbf{25} & \textbf{21} & 7  & 17 \\
{\textbf{Redundancy}}        & 5  & 2  & 5  & 2  & \textbf{13} & \textbf{12} & \textbf{11} & \textbf{11} & 9  & 7  & 8  & 3  & 9  & 5  & 1  & 3  & 8  & 10 & 1  & 7  \\
\hdashline
{\textbf{No error}} & 28 & \textbf{35} & 30 & \textbf{34} & 16 & 20 & 17 & 24 & \textbf{31} & 25 & 28 & \textbf{31} & 13 & 10 & 6  & 20 & 16 & 17 & 11 & 26 \\
\bottomrule
\end{tabular}
\caption{Error analysis on different models and different language scenarios. IB denotes IndicBART and mB denotes to mBART.}
\label{tab:error_stats}
\end{table*}

\subsection{Experimental results and discussions}
\label{sec:results}

\subsubsection{Monolingual}

We list monolingual results in Table~\ref{tab:mono_results2}. Comparing the two extractive baselines, the oracle performance is better than using the first sentence, yet both are nowhere near perfect scores. This indicates that the summary information is scattered across a document and the headline is abstractive in nature.

Our PLM fine-tuning yields significantly higher numbers than extractions, implying a non-trivial headline summarization task. Generally, mBART is ahead of IndicBART, but it supports fewer languages. In terms of automatic metrics, most languages have reasonable performance except for Telugu. The score lags behind perhaps probably due to linguistic features, vocabulary, etc, but not the data quality because we also found that Telugu summarization naturally has a lower score than other languages in prior work \cite{dabre-etal-2022-indicbart}. 

The performance of monolingual English summarization from prompting is listed in Table~\ref{tab:llm_results} along with fine-tuning baselines. Both Vicuna and Alpaca prompting underperform fine-tuning. Also, we discover that although the LLMs can generate some Indian language texts, the quality is subpar, potentially due to the lack of exposure to these languages during pre-training. We hence suggest that our benchmark cannot be trivially solved by LLM prompting because of language inadequacy and domain specificity. Further research is required to overcome the problems.

\subsubsection{Cross-lingual}
Fine-tuning all 182 cross-lingual directions is infeasible given our resource constraint. Thus, we shortlist language pairs to fulfil as much as possible: 1) high and low data availability, 2) combinations of language families as in Table~\ref{tab:language-families}, and 3) languages supported by both IndicBART and mBART for comparison.

According to results in Table~\ref{tab:cross_results}, we observe that summarization-and-translation outperforms fine-tuning, but the gaps are not wide. It is worth noting that the comparison is not strictly fair because the summarization-and-translation pipeline uses an external translation model which presumably ingests much more data. Specifically, the order of translation and summarization matters in higher-resource scenarios: for Hindi and English, translation-summarization achieves 10 more points on all three metrics than summarization-translation. The zero-shot columns reveal that both PLMs are unable to perform cross-lingual summarization even though they have been fine-tuned on monolingual data in all languages. This observation indicates the necessity of our true parallel for practical cross-lingual headline summarization between languages in India.

\subsubsection{Multilingual}
We fine-tune both PLMs on all data where the languages are supported. Results are documented in Table~\ref{tab:multilingual-indicbart} for IndicBART and Table~\ref{tab:multilingual-mbart} for mBART. Regarding cross-lingual cases, multilingual IndicBART seems to only produce sensible numbers for summarization into \texttt{hi}, \texttt{pa}, and \texttt{en}; mBART performs remarkably better than IndicBART. 

Referring to the differences in results ($\Delta$) in the monolingual result Table~\ref{tab:mono_results2}, multilingual models are inferior to monolingual models for both IndicBART and mBART. In clear contrast, for 11 out of 12 cross-lingual directions in Table~\ref{tab:cross_results}, multilingual mBART surpasses separate mBART models fine-tuned for each direction, implying that the availability of data in multiple language pairs helps cross-lingual summarization.

\section{Analysis and Discussions}
\subsection{Error analysis}
To assess and interpret errors made by different models in different language settings, we conducted an error analysis by comparing system outputs with references. We establish error categories and invite human annotators to label the errors in outputs, with setup and explanations of error categories detailed in Appendix~\ref{appendix:error_analysis}. We randomly sample 50 test outputs from all models that deal with monolingual \texttt{en}, \texttt{hi}, \texttt{te}, as well as \texttt{en}$\leftrightarrow$\texttt{hi} cross-lingual summarization. Annotation outcome is presented in Table~\ref{tab:error_stats}.

In monolingual cases, the prevalent error is omission for both IndicBART and mBART models. Conversely, we have different discoveries in cross-lingual models. Dedicated cross-lingual models for a single language pair usually make factuality and omission mistakes. Regarding multilingual systems, IndicBART suffers from language mismatch badly, whereas mBART is associated with factuality, omission and redundancy. Approximately 53\% monolingual summaries and 30\% cross-lingual summaries are considered to be correct. Although decent automatic scores were seen previously, manual inspection suggests that there is ample room for model improvement on our dataset.

\subsection{Languages and PLMs}

\paragraph{Resource availability.} In comparison with Indian languages, we see superior summarization scores for English in the monolingual case, probably due to a larger training size. Multilingual models, too, performed better for languages that have more monolingual data on the target side, for instance, \texttt{en}, \texttt{hi}, and \texttt{gu}.

\paragraph{Unseen language.} Even though Manipuri was not seen during the pre-training of either PLM, it has results close to other languages in monolingual summarization, as well as with multilingual mBART. We hypothesize that this is because the language is written in Bengali script as introduced earlier.

\paragraph{Language family.} In agreement with previous work \cite{hasan-etal-2021-xl}, both IndicBART and mBART work better for Indo-Aryan languages compared to other families.

\paragraph{IndicBART versus mBART.} In monolingual and cross-lingual settings, IndicBART is only slightly behind mBART, being only one-third of the size of mBART. However, for multilingual summarization, mBART is clearly preferred in our context.

\begin{sidewaystable}[htb]
\vspace{50ex}
\centering\scriptsize
\setlength{\tabcolsep}{1ex}
\begin{tabular}{|r|rrr|rrr|rrr|rrr|rrr|rrr|rrr|rrr|}
\hline
& \multicolumn{6}{c|}{\textbf{Summarize-then-translate}} & \multicolumn{6}{c|}{\textbf{Translate-then-summarize}} & \multicolumn{6}{c|}{\textbf{Fine-tuning}}  & \multicolumn{6}{c|}{\textbf{Zero-shot}}\\
 \cline{2-25}
& \multicolumn{3}{c|}{IndicBART} & \multicolumn{3}{c|}{mBART} & \multicolumn{3}{c|}{IndicBART} & \multicolumn{3}{c|}{mBART} & \multicolumn{3}{c|}{IndicBART} & \multicolumn{3}{c|}{mBART} & \multicolumn{3}{c|}{IndicBART} & \multicolumn{3}{c|}{mBART} \\
 & \multicolumn{1}{c}{R2} & \multicolumn{1}{c}{RL} & \multicolumn{1}{c|}{BL} & \multicolumn{1}{c}{R2} & \multicolumn{1}{c}{RL} & \multicolumn{1}{c|}{BL} & \multicolumn{1}{c}{R2} & \multicolumn{1}{c}{RL} & \multicolumn{1}{c|}{BL} & \multicolumn{1}{c}{R2} & \multicolumn{1}{c}{RL} & \multicolumn{1}{c|}{BL} & \multicolumn{1}{c}{R2 ($\Delta$)} & \multicolumn{1}{c}{RL ($\Delta$)} & \multicolumn{1}{c|}{BL ($\Delta$) } & \multicolumn{1}{c}{R2 ($\Delta$)} & \multicolumn{1}{c}{RL ($\Delta$)} & \multicolumn{1}{c|}{BL ($\Delta$) } & \multicolumn{1}{c}{R2} & \multicolumn{1}{c}{RL} & \multicolumn{1}{c|}{BL } & \multicolumn{1}{c}{R2} & \multicolumn{1}{c}{RL} & \multicolumn{1}{c|}{BL}\\
 \hline

\textbf{\texttt{hi-en}} & 24.9 & 41.6 & 5.8 & 25.5 & 43.1 & 6.3 & 40.5 & \textbf{59.8} & 27.0 & 36.3 & 56.8 & 15.8 & \textbf{40.6} (24.1$\uparrow$) & 58.7 (34.7$\uparrow$) & \textbf{28.8} (18.1$\uparrow$) & 35.3 \phantom{0}(8.6$\downarrow$) & 54.9 \phantom{0}(7.2$\downarrow$) & 24.1 \phantom{0}(7.3$\downarrow$) & 0.0& 1.0 & 0.1 & 0.3 & 3.5 & 0.2\\
\textbf{\texttt{en-hi}} & 24.3 & 39.6 & 16.9 & 26.7 & 41.7 & 18.7 & 38.8 & 56.5 & 26.6 & \textbf{39.5} & \textbf{57.5} & \textbf{27.4} & 33.9 (23.1$\uparrow$) & 51.3 (36.0$\uparrow$) & 23.2 (16.7$\uparrow$) & 37.9 \phantom{0}(0.2$\downarrow$) & 55.6 \phantom{0}(1.3$\downarrow$) & 26.8 \phantom{0}(0.5$\uparrow$) & 0.0 & 1.0 & 0.1 & 0.0 & 1.0 & 0.1\\
\textbf{\texttt{gu-te}} & \phantom{0}6.1 & 19.5 & 1.6 & \phantom{0}6.4 & 19.6 & 1.8 & \textbf{16.4} & \textbf{32.8} & \textbf{4.1} & \phantom{0}9.1 & 24.9 & 2.5 & 14.3 (10.5$\uparrow$) & 30.4 (20.7$\uparrow$) & \phantom{0}3.3 \phantom{0}(2.8$\uparrow$) & 14.3 \phantom{0}(1.2$\uparrow$) & 30.3 \phantom{0}(0.8$\uparrow$) & \phantom{0}2.9 \phantom{0}(0.5$\downarrow$) & 0.1 & 1.1 & 0.1 & 0.0 & 0.9 & 0.1 \\
\textbf{\texttt{te-gu}} & \phantom{0}4.4 & 14.6 & 2.1 & \phantom{0}4.4 & 14.8 & 2.0 & 18.0 & 36.6 & \textbf{12.4} & \phantom{0}8.7 & 22.5 & 5.2 & 16.9 (15.4$\uparrow$) & 37.2 (33.4$\uparrow$) & 11.6 (11.2$\uparrow$) & \textbf{20.4} \phantom{0}(1.2$\downarrow$) & \textbf{40.7} \phantom{0}(2.0$\downarrow$) & \textbf{12.4} \phantom{0}(1.3$\downarrow$) & 0.1 & 1.0 & 0.0 & 0.1 & 1.2 & 0.1\\
\textbf{\texttt{ml-mni}} & \textbf{14.9} & \textbf{24.7} & \textbf{12.5} & 13.5 & 23.7 & 9.6 & \phantom{0}8.2 & 16.2 & 5.5 & \phantom{0}8.8 & 16.4 & 6.7 & \phantom{0}8.2 \phantom{0}(2.0$\uparrow$) & 18.9 \phantom{0}(4.2$\uparrow$)& \phantom{0}4.8 \phantom{0}(1.5$\uparrow$) & \phantom{0}7.6 \phantom{0}(8.3$\downarrow$) & 18.9 (12.5$\downarrow$) & \phantom{0}2.5 \phantom{0}(6.4$\downarrow$) & 0.0 & 0.0 & 0.1 & 0.0 & 0.0 & 0.1 \\
\textbf{\texttt{mni-ml}} & \phantom{0}\textbf{7.5} & \textbf{22.3} & \textbf{4.4} & \phantom{0}7.4 & 22.1 & 3.9 & \phantom{0}5.8 & 19.4 & 1.3 & \phantom{0}3.3 & 14.7 & 0.9 & \phantom{0}4.7 \phantom{0}(4.7$\uparrow$) & 14.6 (14.5$\uparrow$)& \phantom{0}1.9 (\phantom{0}1.9$\uparrow$) & \phantom{0}0.6 (10.4$\downarrow$) & \phantom{0}6.8 (21.1$\downarrow$) & \phantom{0}0.5 \phantom{0}(4.6$\downarrow$) & 0.0 & 0.1 & 0.0 & 0.0 & 0.0 & 0.1  \\
\textbf{\texttt{mr-bn}} & 13.9 & 32.2 & 8.7 & \textbf{14.1} & \textbf{33.0} & \textbf{9.5} & 12.7 & 31.7 & 7.1 & 11.5 & 31.2 & 7.7 & 12.9 (10.4$\uparrow$) & 30.4 (23.2$\uparrow$) & \phantom{0}7.2 \phantom{0}(6.3$\uparrow$) & 12.8 \phantom{0}(2.8$\downarrow$) & 31.5 \phantom{0}(3.0$\downarrow$) & \phantom{0}7.0 \phantom{0}(3.2$\downarrow$) & 0.0 & 0.0 & 0.0 & 0.0 & 0.0 & 0.1\\
\textbf{\texttt{bn-mr}} & 13.4 & 31.3 & 8.6 & 14.4 & 32.8 & 9.1 & \textbf{19.1} & 36.7 & \textbf{11.1} & 17.8 & 34.9 & \textbf{11.1} & 16.6 (13.4$\uparrow$) & 35.5 (30.0$\uparrow$) & \phantom{0}9.7 \phantom{0}(8.6$\uparrow$) & 19.1 \phantom{0}(1.3$\downarrow$) & \textbf{37.1} \phantom{0}(1.9$\downarrow$) & 10.5 \phantom{0}(1.7$\downarrow$) & 0.0 & 0.0 & 0.0 & 0.0 & 0.0 & 0.0 \\
\textbf{\texttt{te-ta}} & 14.3 & 30.5 & 9.5 & 13.5 & 31.1 & 9.5 & 24.3 & 41.5 & 19.2 & \textbf{25.1} & \textbf{43.0} & \textbf{20.3} & 17.9 \phantom{0}(4.6$\uparrow$) & 34.8 \phantom{0}(5.1$\uparrow$) & 11.4 \phantom{0}(3.4$\uparrow$) & 16.5 \phantom{0}(2.6$\downarrow$) & 35.7 \phantom{0}(4.0$\downarrow$) & \phantom{0}8.2 \phantom{0}(3.4$\downarrow$) & 0.1 & 1.1 & 0.0 & 0.1 & 1.2 & 0.0 \\
\textbf{\texttt{ta-te}} & \textbf{22.5} & 37.8 & 13.4 & 22.3 & \textbf{38.4} & \textbf{13.6} & 14.0 & 29.6 & 4.8 & 13.2 & 30.1 & 5.3 & 15.7 \phantom{0}(7.9$\uparrow$) & 30.9 (15.0$\uparrow$) & \phantom{0}2.5 \phantom{0}(1.2$\uparrow$) & 14.2 \phantom{0}(0.2$\downarrow$) & 29.4 \phantom{0}(0.4$\downarrow$) & \phantom{0}2.8 \phantom{0}(0.5$\downarrow$) & 0.1& 1.4 & 0.1 & 0.1 & 1.3 & 0.1 \\
\textbf{\texttt{mni-en}} & 15.4 & 28.4 & 2.9 & 16.7 & 30.0 & 4.0 & \textbf{19.3} & \textbf{38.1} & 11.5 & 14.5 & 31.6 & 7.6 & 18.1 (16.4$\uparrow$) & 35.7 (33.3$\uparrow$) & \textbf{12.3} (11.3$\uparrow$) & \phantom{0}9.3 (28.8$\downarrow$) & 22.2 (35.2$\downarrow$) & \phantom{0}7.2 (19.8$\downarrow$) & 0.0& 0.1 & 0.0 & 0.2 & 1.6 & 0.1 \\
\textbf{\texttt{en-mni}} & \textbf{24.0} & \textbf{35.7} & \textbf{18.2} & 23.7 & 35.3 & 18.0 & 12.9 & 22.0 & 9.5 & 14.2 & 23.9 & 11.3 & \phantom{0}7.8 \phantom{0}(0.6$\uparrow$) & 18.4 \phantom{0}(1.7$\uparrow$) & \phantom{0}3.6 \phantom{0}(0.1$\downarrow$) & \phantom{0}5.9 (11.0$\downarrow$) & 14.9 (18.4$\downarrow$) & \phantom{0}1.5 \phantom{0}(8.6$\downarrow$) & 0.0 & 0.0 & 0.1 & 0.0 & 0.0 & 0.1 \\

\hline
\end{tabular}
\vspace{-1ex}
\caption{Cross-lingual benchmarks: separate models for each language pair. Bold indicates the best result; ($\Delta$) refers to the change relative to the corresponding multilingual result in Table~\ref{tab:multilingual-indicbart} or Table~\ref{tab:multilingual-mbart}.}
\label{tab:cross_results}
\vspace{6ex}
\setlength{\tabcolsep}{0.75ex}
\centering\scriptsize
\begin{tabular}{|r|rrr|rrr|rrr|rrr|rrr|rrr|rrr|rrr|rrr|rrr|rrr|rrr|rrr|} 
 \hline
& \multicolumn{3}{c|}{\textbf{\texttt{as}}} & \multicolumn{3}{c|}{\textbf{\texttt{bn}}} & \multicolumn{3}{c|}{\textbf{\texttt{gu}}} & \multicolumn{3}{c|}{\textbf{\texttt{\texttt{hi}}}} & \multicolumn{3}{c|}{\textbf{\texttt{kn}}} & \multicolumn{3}{c|}{\textbf{\texttt{ml}}} & \multicolumn{3}{c|}{\textbf{\texttt{mni}}} & \multicolumn{3}{c|}{\textbf{\texttt{mr}}} & \multicolumn{3}{c|}{\textbf{\texttt{or}}} & \multicolumn{3}{c|}{\textbf{\texttt{pa}}} & \multicolumn{3}{c|}{\textbf{\texttt{ta}}} & \multicolumn{3}{c|}{\textbf{\texttt{te}}} & \multicolumn{3}{c|}{\textbf{\texttt{en}}} \\
 & \multicolumn{1}{c}{R2} & \multicolumn{1}{c}{RL} & \multicolumn{1}{c|}{BL} & \multicolumn{1}{c}{R2} & \multicolumn{1}{c}{RL} & \multicolumn{1}{c|}{BL} & \multicolumn{1}{c}{R2} & \multicolumn{1}{c}{RL} & \multicolumn{1}{c|}{BL} & \multicolumn{1}{c}{R2} & \multicolumn{1}{c}{RL} & \multicolumn{1}{c|}{BL} & \multicolumn{1}{c}{R2} & \multicolumn{1}{c}{RL} & \multicolumn{1}{c|}{BL} & \multicolumn{1}{c}{R2} & \multicolumn{1}{c}{RL} & \multicolumn{1}{c|}{BL} & \multicolumn{1}{c}{R2} & \multicolumn{1}{c}{RL} & \multicolumn{1}{c|}{BL} & \multicolumn{1}{c}{R2} & \multicolumn{1}{c}{RL} & \multicolumn{1}{c|}{BL} & \multicolumn{1}{c}{R2} & \multicolumn{1}{c}{RL} & \multicolumn{1}{c|}{BL} & \multicolumn{1}{c}{R2} & \multicolumn{1}{c}{RL} & \multicolumn{1}{c|}{BL} & \multicolumn{1}{c}{R2} & \multicolumn{1}{c}{RL} & \multicolumn{1}{c|}{BL} & \multicolumn{1}{c}{R2} & \multicolumn{1}{c}{RL} & \multicolumn{1}{c|}{BL} & \multicolumn{1}{c}{R2} & \multicolumn{1}{c}{RL} & \multicolumn{1}{c|}{BL} \\
 \hline

\textbf{\texttt{as}} & 26.3 & 37.6 & 17.7 & 3.0 & 9.9 & 1.2 & 7.3 & 10.7 & 3.4 & 9.0 & 12.5 & 5.9 & 3.3 & 6.5 & 2.4 & 1.6 & 2.9 & 0.1 & 10.0 & 23.4 & 4.9 & 2.3 & 4.2 & 1.1 & 5.0 & 7.6 & 1.8 & 10.6 & 16.1 & 6.4 & 0.6 & 1.6 & 0.3 & 4.2 & 9.7 & 0.5 & 12.8 & 18.6 & 7.9 \\
\textbf{\texttt{bn}} & 2.3 & 7.7 & 0.7 & 17.3 & 29.2 & 10.4 & 6.2 & 11.0 & 3.0 & 6.1 & 9.9 & 3.8 & 4.4 & 7.4 & 2.2 & 1.0 & 2.4 & 0.6 & 10.3 & 23.2 & 5.4 & 3.2 & 5.5 & 1.1 & 5.3 & 9.0 & 2.0 & 10.8 & 16.5 & 6.3 & 1.1 & 2.0 & 0.3 & 3.4 & 9.0 & 0.8 & 14.1 & 21.2 & 9.0 \\
\textbf{\texttt{gu}} & 0.4 & 2.5 & 0.2 & 1.3 & 4.3 & 0.3 & 45.4 & 57.9 & 36.6 & 3.9 & 6.4 & 3.2 & 2.4 & 5.0 & 1.2 & 1.0 & 2.5 & 0.3 & 5.1 & 11.9 & 2.7 & 3.3 & 6.1 & 1.1 & 2.1 & 4.4 & 0.7 & 5.1 & 8.1 & 4.2 & 2.4 & 4.7 & 0.7 & 3.8 & 9.7 & 0.5 & 10.0 & 14.6 & 7.8\\
\textbf{\texttt{hi}} & 1.7 & 5.5 & 0.7 & 0.9 & 4.0 & 0.2 & 5.0 & 9.5 & 1.9 & 48.5 & 60.6 & 38.7 & 5.7 & 10.4 & 3.2 & 2.8 & 6.3 & 0.7 & 7.0 & 16.2 & 3.9 & 6.6 & 13.8 & 2.6 & 5.4 & 9.3 & 1.7 & 17.2 & 25.2 & 11.2 & 4.1 & 7.5 & 1.7 & 6.4 & 14.8 & 0.8 & 16.5 & 24.0 & 10.7\\
\textbf{\texttt{kn}} & 1.6 & 4.5 & 0.4 & 1.6 & 5.1 & 0.4 & 7.7 & 14.4 & 4.1 & 9.3 & 13.8 & 6.3 & 35.6 & 48.5 & 28.5 & 1.7 & 3.4 & 0.3 & 7.1 & 15.7 & 3.4 & 2.8 & 5.5 & 1.1 & 3.9 & 7.2 & 1.5 & 14.1 & 20.9 & 8.7 & 3.5 & 5.9 & 1.3 & 6.9 & 16.7 & 1.0 & 16.8 & 26.6 & 10.3\\
\textbf{\texttt{ml}} & 1.5 & 4.7 & 0.5 & 1.3 & 4.8 & 0.4 & 9.3 & 17.0 & 4.3 & 11.4 & 17.6 & 6.3 & 6.4 & 14.0 & 3.7 & 24.0 & 40.0 & 10.2 & 6.2 & 14.7 & 3.3 & 3.8 & 8.2 & 1.5 & 5.7 & 10.7 & 1.9 & 19.1 & 27.9 & 12.9 & 4.7 & 9.6 & 1.4 & 9.7 & 21.3 & 1.9 & 20.2 & 29.8 & 13.4\\
\textbf{\texttt{mni}} & 0.5 & 3.0 & 0.1 & 0.5 & 4.3 & 0.3 & 0.2 & 0.5 & 0.1 & 0.4 & 0.5 & 0.3 & 0.9 & 1.2 & 0.5 & 0.0 & 0.1 & 0.0 & 39.4 & 54.2 & 32.8 & 0.3 & 0.4 & 0.1 & 1.4 & 2.2 & 0.5 & 0.6 & 0.9 & 0.6 & 0.0 & 0.3 & 0.0 & 0.6 & 1.6 & 0.1 & 1.7 & 2.4 & 1.0\\
\textbf{\texttt{mr}}  & 1.9 & 5.2 & 0.6 & 2.5 & 7.2 & 0.9 & 11.6 & 19.5 & 4.9 & 21.9 & 33.6 & 13.6 & 6.4 & 11.7 & 3.0 & 4.5 & 9.5 & 0.9 & 7.8 & 17.5 & 3.8 & 32.4 & 45.2 & 25.0 & 8.0 & 13.5 & 3.0 & 19.3 & 27.9 & 13.1 & 3.4 & 7.0 & 1.4 & 7.1 & 15.4 & 0.8 & 20.0 & 28.9 & 12.9 \\
\textbf{\texttt{or}} &  2.5 & 5.9 & 0.9 & 2.8 & 7.5 & 0.8 & 8.6 & 15.6 & 4.3 & 11.5 & 17.8 & 8.5 & 4.0 & 8.0 & 2.0 & 2.2 & 4.9 & 0.4 & 8.1 & 18.9 & 5.1 & 5.2 & 8.8 & 1.6 & 35.6 & 49.2 & 23.6 & 15.8 & 22.8 & 11.5 & 3.8 & 6.2 & 1.4 & 6.7 & 13.9 & 0.8 & 14.9 & 22.8 & 8.6\\
\textbf{\texttt{pa}} & 0.7 & 2.9 & 0.3 & 1.7 & 4.7 & 0.4 & 5.1 & 9.1 & 1.7 & 11.3 & 16.0 & 7.6 & 4.9 & 8.8 & 2.4 & 1.7 & 3.4 & 0.7 & 5.8 & 14.1 & 2.6 & 5.2 & 8.8 & 2.0 & 3.1 & 5.5 & 0.9 & 50.8 & 64.2 & 43.0 & 2.2 & 3.5 & 0.9 & 6.7 & 14.4 & 0.5 & 13.5 & 19.9 & 7.6 \\
\textbf{\texttt{ta}}&0.8 &2.6 &0.3 &1.7 &4.9 &0.4 &9.5 &16.8 &4.9 &11.0&16.7&7.1&4.7&8.2&2.0&2.1&5.1&0.7&6.4&14.8&3.2&2.7&5.3&1.0&4.4&7.2&1.5&11.7&17.0&7.4&41.0&56.4&31.6&7.8&15.9&1.3&17.1&24.5&11.7 \\
\textbf{\texttt{te}} & 0.2&1.0&0.0&0.5&1.9&0.2&1.5&3.7&0.4&2.2&3.5&1.0&0.4&1.4&0.1&0.2&0.9&0.0&2.8&6.6&1.3&0.1&0.9&0.1&0.1&1.1&0.1&3.0&4.8&1.4&0.4&1.0&0.1&13.3&29.7&8.0&4.9&7.8&2.1 \\
\textbf{\texttt{en}} &  0.6&2.1&0.2&1.1&5.0&0.2&7.7&13.7&3.0&10.8&15.3&6.5&5.5&10.4&2.8&2.1&4.3&0.2&7.2&16.7&3.6&3.8&8.2&1.6&3.1&6.0&0.9&12.2&18.4&7.3&4.0&6.9&1.2&6.7&14.7&0.8&57.9&69.2&45.4\\

\hline
\end{tabular}
\vspace{-1ex}
\caption{Multilingual benchmark with IndicBART: a single model for $13\times{}13=169$ supported language pairs.}
\label{tab:multilingual-indicbart}

\vspace{6ex}

\setlength{\tabcolsep}{1ex}
\centering\scriptsize
\begin{tabular}{|r|rrr|rrr|rrr|rrr|rrr|rrr|rrr|rrr|rrr|rrr|} 
 \hline
& \multicolumn{3}{c|}{\textbf{\texttt{bn}}} & \multicolumn{3}{c|}{\textbf{\texttt{gu}}} & \multicolumn{3}{c|}{\textbf{\texttt{hi}}} & \multicolumn{3}{c|}{\textbf{\texttt{ml}}} & \multicolumn{3}{c|}{\textbf{\texttt{mni}}} & \multicolumn{3}{c|}{\textbf{\texttt{mr}}} & \multicolumn{3}{c|}{\textbf{\texttt{ta}}} & \multicolumn{3}{c|}{\textbf{\texttt{te}}} & \multicolumn{3}{c|}{\textbf{\texttt{ur}}} & \multicolumn{3}{c|}{\textbf{\texttt{en}}} \\
 & \multicolumn{1}{c}{R2} & \multicolumn{1}{c}{RL} & \multicolumn{1}{c|}{BL} & \multicolumn{1}{c}{R2} & \multicolumn{1}{c}{RL} & \multicolumn{1}{c|}{BL} & \multicolumn{1}{c}{R2} & \multicolumn{1}{c}{RL} & \multicolumn{1}{c|}{BL} & \multicolumn{1}{c}{R2} & \multicolumn{1}{c}{RL} & \multicolumn{1}{c|}{BL} & \multicolumn{1}{c}{R2} & \multicolumn{1}{c}{RL} & \multicolumn{1}{c|}{BL} & \multicolumn{1}{c}{R2} & \multicolumn{1}{c}{RL} & \multicolumn{1}{c|}{BL} & \multicolumn{1}{c}{R2} & \multicolumn{1}{c}{RL} & \multicolumn{1}{c|}{BL} & \multicolumn{1}{c}{R2} & \multicolumn{1}{c}{RL} & \multicolumn{1}{c|}{BL} & \multicolumn{1}{c}{R2} & \multicolumn{1}{c}{RL} & \multicolumn{1}{c|}{BL} & \multicolumn{1}{c}{R2} & \multicolumn{1}{c}{RL} & \multicolumn{1}{c|}{BL} \\
\hline

\textbf{\texttt{bn}} & 29.2 & 47.9 & 20.4 & 23.5 & 44.4 & 14.6 & 34.3 & 53.9 & 22.9 & 14.1 & 31.7 & 6.9 & 15.4 & 31.7 & 8.5 & 20.4 & 39.0 & 12.2 & 18.7 & 39.1 & 10.7 & 12.6 & 28.5 & 3.5 & 27.9 & 45.6 & 19.3 & 39.8 & 59.1 & 28.9 \\
\textbf{\texttt{gu}} & 16.4 & 35.1 & 9.7 & 48.3 & 65.1 & 41.2 & 36.6 & 55.3 & 25.2 & 14.5 & 31.1 & 6.9 & 15.6 & 31.4 & 8.8 & 21.3 & 39.8 & 13.5 & 19.1 & 39.1 & 11.3 & 13.1 & 29.5 & 3.4 & 28.2 & 45.8 & 20.0 & 42.9 & 61.0 & 30.3\\
\textbf{\texttt{hi}} &15.9 & 33.9 & 9.7 & 24.8 & 46.0 & 15.6 & 54.5 & 69.0 & 47.3 & 14.0 & 31.4 & 6.4 & 15.7 & 31.8 & 9.3 & 22.4 & 40.8 & 14.9 & 19.8 & 40.4 & 11.2 & 15.0 & 31.7 & 3.7 & 28.7 & 46.5 & 20.4 & 43.9 & 62.1 & 31.4\\
\textbf{\texttt{ml}} & 15.4 & 34.7 & 9.3 & 22.7 & 43.4 & 13.9 & 36.2 & 54.8 & 24.6 & 27.1 & 45.6 & 15.8 & 15.9 & 31.4 & 8.9 & 19.6 & 38.4 & 12.3 & 18.6 & 38.5 & 10.4 & 13.6 & 29.6 & 3.6 & 27.3 & 45.2 & 19.0 & 41.3 & 60.5 & 29.1\\
\textbf{\texttt{mni}} & 12.0 & 30.3 & 7.1 & 21.2 & 41.3 & 13.6 & 32.3 & 51.3 & 21.9 & 11.0 & 27.9 & 5.1 & 39.8 & 55.5 & 33.4 & 17.8 & 36.5 & 11.0 & 17.1 & 36.1 & 8.6 & 12.3 & 27.9 & 2.8 & 25.9 & 43.1 & 18.2 & 38.1 & 57.4 & 27.0\\
\textbf{\texttt{mr}} & 15.6 & 34.5 & 10.2 & 24.5 & 45.7 & 15.1 & 38.2 & 57.0 & 26.3 & 14.4 & 31.2 & 6.4 & 15.6 & 31.4 & 8.4 & 40.2 & 58.0 & 35.8 & 18.8 & 39.5 & 10.7 & 13.7 & 30.0 & 3.9 & 26.7 & 44.5 & 18.5 & 41.9 & 61.0 & 30.3\\
\textbf{\texttt{ta}} &15.2 & 33.0 & 8.4 & 21.2 & 41.9 & 12.6 & 34.0 & 52.9 & 23.1 & 13.1 & 29.8 & 6.0 & 13.4 & 29.1 & 6.2 & 19.6 & 37.8 & 12.5 & 41.6 & 59.2 & 35.1 & 14.4 & 29.8 & 3.2 & 26.4 & 44.2 & 18.0 & 40.7 & 59.7 & 29.1  \\
\textbf{\texttt{te}} & 15.1 & 32.9 & 9.4 & 21.6 & 42.7 & 13.7 & 36.1 & 55.1 & 23.5 & 13.1 & 30.2 & 6.6 & 16.2 & 31.7 & 9.3 & 21.1 & 39.5 & 13.7 & 19.1 & 39.7 & 11.6 & 15.6 & 33.4 & 9.0 & 26.7 & 44.7 & 18.4 & 40.5 & 59.7 & 29.5 \\
\textbf{\texttt{ur}} &14.5 & 32.8 & 8.8 & 23.8 & 44.6 & 15.3 & 34.7 & 54.0 & 23.0 & 12.7 & 29.5 & 6.2 & 14.5 & 30.6 & 7.7 & 20.3 & 39.0 & 12.1 & 19.1 & 39.6 & 11.2 & 14.3 & 29.9 & 3.2 & 48.3 & 62.2 & 40.8 & 41.9 & 60.3 & 30.6 \\
\textbf{\texttt{en}} & 15.7 & 34.3 & 9.5 & 25.1 & 46.4 & 15.5 & 38.1 & 56.9 & 26.3 & 13.6 & 31.1 & 6.0 & 16.9 & 33.3 & 10.1 & 20.7 & 39.4 & 13.4 & 20.2 & 40.6 & 11.9 & 14.3 & 30.5 & 4.0 & 29.9 & 48.0 & 21.3 & 64.0 & 76.7 & 51.5\\

\hline
\end{tabular}
\vspace{-1ex}
\caption{Multilingual benchmark with mBART: a single model for $10\times{}10=100$ supported language pairs.}
\label{tab:multilingual-mbart}
\end{sidewaystable}

\FloatBarrier
\section{Related Work}\label{sec:related_works}

Document-summary pairs are the key components of a summarization dataset. A typical acquisition workflow is to utilize publicly available articles, given their high availability. To form a document-summary pair, an article can be paired with either human-written summaries, or accompanying texts such as the headline or the first sentence.

\subsection{Data acquisition}
Earlier works like the CNN/DM corpus utilized English news articles as documents and highlights as summaries \citep{hermann-etal-2015-teaching,nallapati-etal-2016-abstractive}, followed by \citet{scialom-etal-2020-mlsum2} to construct MLSUM in five Indo-European languages. \citet{hasan-etal-2021-xl} created XL-Sum with BBC articles' first sentence as a summary and the rest as a document. \citet{varab-schluter-2021-massivesumm} developed MassiveSumm by scraping a wide range of websites. The ILSUM dataset crawled several leading newspapers in India \cite{satapara2022fire}, and TeSum used Telugu news with summaries from human annotators \citep{urlana-etal-2022-tesum}. \citet{verma-etal-2023-large} built a multilingual multi-modal dataset M3LS from the BBC website. Recently, a concurrent work sourced data from a news aggregator to form a multilingual headline dataset in Indic languages \citep[V\=arta,][]{aralikatte2023varta}.

Exploiting public resources usually results in same-language document-summary pairs as introduced above. 
\citet{zhu-etal-2019-ncls2} translated summaries in monolingual datasets into another language while ensuring that the round-trip translation of the summary did not deviate from the original one. The CrossSum corpus \citep{Bhattacharjee2022crosssum} was derived from XL-Sum by thresholding similarity between cross-lingual summaries, and pairing a summary with the document of a parallel summary. The naturally arising cross-lingual datasets include WikiLingua which aligned summaries and documents using image pivots \citep{ladhak-etal-2020-wikilingua} and EUR-Lex-Sum from a platform hosting legal documents \citep{aumiller-etal-2022-eur}.

Our PMIndiaSum falls in the category of using public article-headline pairs. The data itself is massively cross-lingual due to the multilingualism of the published news. We greatly benefit from the materials released by the PMIndia parallel corpus \cite{haddow2020pmindia}, which is a machine translation dataset between English and Indian languages, extracted from the same website.

\subsection{Coverage for languages in India}
The current availability of Indian language summariztaion data is limited. TeSum is monolingual, ILSUM covers three Indian languages, XL-Sum supports eight, and M3LS covers ten. Covering the largest number of Indian languages are MassiveSumm and V\=arta, but these works release URLs and processing scripts instead of the data; copyright holders' conditions and licenses remain unknown. Such practices transfer the risk and responsibility to dataset users. Our data can be freely used, modified, and re-distributed.

Moreover, the above multilingual datasets cannot be used for cross-lingual research. Existing data for summarization between Indian languages are scarce. WikiLingua supports one Indian language, and to the best of our knowledge, before ours, CrossSum is the largest which contains 56 language pairs from pairing up 8 languages in XL-Sum. In comparison, even excluding English, our PMIndiaSum contains 13 Indian languages and 156 cross-lingual pairs. In a zero-shot fashion, one can train on multilingual datasets for cross-lingual capability. Nonetheless, the performance may be inferior as we have shown. More importantly, monolingual and multilingual datasets cannot provide a benchmark to evaluate cross-lingual research. Our work fills the gap by providing both training data and a trustworthy test suite for a large number of Indian language directions.

\section{Conclusion and Future Work}
We have presented PMIndiaSum, a headline summarization corpus focused on 14 languages in India, supporting monolingual, cross-lingual, and multilingual summarization in 196 directions. We outlined the processing steps, explained data statistics, discussed quality considerations, and made the dataset publicly accessible. We also published benchmarks for extractive baselines, summarize-and-translate, and fine-tuning, involving two pre-trained language models. Experiment results emphasize the value of our dataset for summarization between languages in India.

A natural extension is to continuously integrate new articles from the source website to update the dataset. However, to ensure that experimental comparisons remain fair, one needs to maintain consistent data sizes in model training. Another direction is to ingest more websites to expand the size, domain coverage, language availability, and modality.

\section*{Limitations}
Our data sources are scraped from a government website, leading to a domain bias towards political news, and a style bias towards headline-like summaries. Also, we selected the articles available in all languages to be validation and testing instances, to prevent information leakage. While we believe this to be a sensible decision, those articles might carry a distributional drift, such as being more important for all readers or easier to translate.

\section*{Ethics Statement}
We place trust in the Indian government website to eliminate inappropriate content, but we could not inspect each data instance ourselves. On the other hand, we note that the values promoted by a governmental agency might not align with every potential user of the data. Also, the provision of data in widely spoken languages might further edge out lesser-used languages in research.

\section*{Acknowledgements}
We thank Dennis Aumiller and the anonymous reviewers for giving feedback. We are grateful to Pruthwik Mishra, Lokesh Madasu, Kalahasti Ganesh Srivatsa, and Nikhilesh Bhatnagar for helping with human evaluation. We also thank Marcio Fonseca for providing the code for LLM prompting, and LTRC, IIIT Hyderabad for providing the translation engine for our summarization-and-translation experiments. Our work would not be possible if the Prime Minister's Office of the Government of India or the authors of the PMIndia parallel corpus did not allow the re-distribution of their materials.

Pinzhen Chen and Barry Haddow are funded by UK Research and Innovation (UKRI) under the UK government’s Horizon Europe funding guarantee [grant numbers 10052546 and 10039436]. Zheng Zhao is supported by the UKRI Centre for Doctoral Training in Natural Language Processing (EP/S022481/1).

\bibliography{custom, anthology}

\begin{thebibliography}{42}
\expandafter\ifx\csname natexlab\endcsname\relax\def\natexlab#1{#1}\fi

\bibitem[{Aralikatte et~al.(2023)Aralikatte, Cheng, Doddapaneni, and
  Cheung}]{aralikatte2023varta}
Rahul Aralikatte, Ziling Cheng, Sumanth Doddapaneni, and Jackie Chi~Kit Cheung.
  2023.
\newblock \href {https://doi.org/10.18653/v1/2023.findings-acl.215} {Varta: A
  large-scale headline-generation dataset for {I}ndic languages}.
\newblock In \emph{Findings of the Association for Computational Linguistics:
  ACL 2023}.

\bibitem[{Aumiller et~al.(2022)Aumiller, Chouhan, and
  Gertz}]{aumiller-etal-2022-eur}
Dennis Aumiller, Ashish Chouhan, and Michael Gertz. 2022.
\newblock \href {https://aclanthology.org/2022.emnlp-main.519} {{EUR}-lex-sum:
  A multi- and cross-lingual dataset for long-form summarization in the legal
  domain}.
\newblock In \emph{Proceedings of the 2022 Conference on Empirical Methods in
  Natural Language Processing}.

\bibitem[{Bhattacharjee et~al.(2023)Bhattacharjee, Hasan, Ahmad, Li, Kang, and
  Shahriyar}]{Bhattacharjee2022crosssum}
Abhik Bhattacharjee, Tahmid Hasan, Wasi~Uddin Ahmad, Yuan-Fang Li, Yong-Bin
  Kang, and Rifat Shahriyar. 2023.
\newblock \href {https://doi.org/10.18653/v1/2023.acl-long.143} {{C}ross{S}um:
  Beyond {E}nglish-centric cross-lingual summarization for 1,500+ language
  pairs}.
\newblock In \emph{Proceedings of the 61st Annual Meeting of the Association
  for Computational Linguistics}.

\bibitem[{Bommasani and Cardie(2020)}]{bommasani-cardie-2020-intrinsic2}
Rishi Bommasani and Claire Cardie. 2020.
\newblock \href {https://doi.org/10.18653/v1/2020.emnlp-main.649} {Intrinsic
  evaluation of summarization datasets}.
\newblock In \emph{Proceedings of the 2020 Conference on Empirical Methods in
  Natural Language Processing}.

\bibitem[{Chiang et~al.(2023)Chiang, Li, Lin, Sheng, Wu, Zhang, Zheng, Zhuang,
  Zhuang, Gonzalez, Stoica, and Xing}]{vicuna2023}
Wei-Lin Chiang, Zhuohan Li, Zi~Lin, Ying Sheng, Zhanghao Wu, Hao Zhang, Lianmin
  Zheng, Siyuan Zhuang, Yonghao Zhuang, Joseph~E. Gonzalez, Ion Stoica, and
  Eric~P. Xing. 2023.
\newblock \href {https://lmsys.org/blog/2023-03-30-vicuna/} {Vicuna: An
  open-source chatbot impressing {GPT-4} with 90\%* {ChatGPT} quality}.
\newblock {lmsys.org}.

\bibitem[{Clark et~al.(2023)Clark, Rijhwani, Gehrmann, Maynez, Aharoni,
  Nikolaev, Sellam, Siddhant, Das, and Parikh}]{clark2023seahorse}
Elizabeth Clark, Shruti Rijhwani, Sebastian Gehrmann, Joshua Maynez, Roee
  Aharoni, Vitaly Nikolaev, Thibault Sellam, Aditya Siddhant, Dipanjan Das, and
  Ankur~P Parikh. 2023.
\newblock \href {https://arxiv.org/abs/2305.13194v1} {{SEAHORSE}: A
  multilingual, multifaceted dataset for summarization evaluation}.
\newblock \emph{arXiv preprint}.

\bibitem[{Dabre et~al.(2022)Dabre, Shrotriya, Kunchukuttan, Puduppully, Khapra,
  and Kumar}]{dabre-etal-2022-indicbart}
Raj Dabre, Himani Shrotriya, Anoop Kunchukuttan, Ratish Puduppully, Mitesh
  Khapra, and Pratyush Kumar. 2022.
\newblock \href {https://doi.org/10.18653/v1/2022.findings-acl.145}
  {{I}ndic{BART}: A pre-trained model for indic natural language generation}.
\newblock In \emph{Findings of the Association for Computational Linguistics:
  ACL 2022}.

\bibitem[{Devlin et~al.(2019)Devlin, Chang, Lee, and
  Toutanova}]{devlin-etal-2019-bert2}
Jacob Devlin, Ming-Wei Chang, Kenton Lee, and Kristina Toutanova. 2019.
\newblock \href {https://doi.org/10.18653/v1/N19-1423} {{BERT}: Pre-training of
  deep bidirectional transformers for language understanding}.
\newblock In \emph{Proceedings of the 2019 Conference of the North {A}merican
  Chapter of the Association for Computational Linguistics: Human Language
  Technologies}.

\bibitem[{Feng et~al.(2022)Feng, Yang, Cer, Arivazhagan, and
  Wang}]{feng-etal-2022-language2}
Fangxiaoyu Feng, Yinfei Yang, Daniel Cer, Naveen Arivazhagan, and Wei Wang.
  2022.
\newblock \href {https://doi.org/10.18653/v1/2022.acl-long.62}
  {Language-agnostic {BERT} sentence embedding}.
\newblock In \emph{Proceedings of the 60th Annual Meeting of the Association
  for Computational Linguistics}.

\bibitem[{Gebru et~al.(2021)Gebru, Morgenstern, Vecchione, Vaughan, Wallach,
  III, and Crawford}]{datasheet}
Timnit Gebru, Jamie Morgenstern, Briana Vecchione, Jennifer~Wortman Vaughan,
  Hanna Wallach, Hal~Daum\'{e} III, and Kate Crawford. 2021.
\newblock \href {https://doi.org/10.1145/3458723} {Datasheets for datasets}.
\newblock \emph{Commun. ACM}, 64(12).

\bibitem[{Grusky et~al.(2018)Grusky, Naaman, and
  Artzi}]{grusky-etal-2018-newsroom2}
Max Grusky, Mor Naaman, and Yoav Artzi. 2018.
\newblock \href {https://doi.org/10.18653/v1/N18-1065} {{N}ewsroom: A dataset
  of 1.3 million summaries with diverse extractive strategies}.
\newblock In \emph{Proceedings of the 2018 Conference of the North {A}merican
  Chapter of the Association for Computational Linguistics: Human Language
  Technologies}.

\bibitem[{Haddow and Kirefu(2020)}]{haddow2020pmindia}
Barry Haddow and Faheem Kirefu. 2020.
\newblock \href {https://arxiv.org/abs/2001.09907v1} {{PMIndia}---{A}
  collection of parallel corpora of languages of {India}}.
\newblock \emph{arXiv preprint}.

\bibitem[{Hasan et~al.(2021)Hasan, Bhattacharjee, Islam, Mubasshir, Li, Kang,
  Rahman, and Shahriyar}]{hasan-etal-2021-xl}
Tahmid Hasan, Abhik Bhattacharjee, Md.~Saiful Islam, Kazi Mubasshir, Yuan-Fang
  Li, Yong-Bin Kang, M.~Sohel Rahman, and Rifat Shahriyar. 2021.
\newblock \href {https://doi.org/10.18653/v1/2021.findings-acl.413} {{XL}-sum:
  Large-scale multilingual abstractive summarization for 44 languages}.
\newblock In \emph{Findings of the Association for Computational Linguistics:
  ACL-IJCNLP 2021}.

\bibitem[{Hermann et~al.(2015)Hermann, Kocisky, Grefenstette, Espeholt, Kay,
  Suleyman, and Blunsom}]{hermann-etal-2015-teaching}
Karl~Moritz Hermann, Tomas Kocisky, Edward Grefenstette, Lasse Espeholt, Will
  Kay, Mustafa Suleyman, and Phil Blunsom. 2015.
\newblock \href
  {https://proceedings.neurips.cc/paper_files/paper/2015/file/afdec7005cc9f14302cd0474fd0f3c96-Paper.pdf}
  {Teaching machines to read and comprehend}.
\newblock In \emph{Advances in Neural Information Processing Systems}.

\bibitem[{Houlsby et~al.(2019)Houlsby, Giurgiu, Jastrzebski, Morrone,
  De~Laroussilhe, Gesmundo, Attariyan, and Gelly}]{pmlr-v97-houlsby19a}
Neil Houlsby, Andrei Giurgiu, Stanislaw Jastrzebski, Bruna Morrone, Quentin
  De~Laroussilhe, Andrea Gesmundo, Mona Attariyan, and Sylvain Gelly. 2019.
\newblock \href {https://proceedings.mlr.press/v97/houlsby19a.html}
  {Parameter-efficient transfer learning for {NLP}}.
\newblock In \emph{Proceedings of the 36th International Conference on Machine
  Learning}.

\bibitem[{Koo and Li(2016)}]{koo2016guideline}
Terry~K Koo and Mae~Y Li. 2016.
\newblock \href {https://doi.org/10.1016/j.jcm.2016.02.012} {A guideline of
  selecting and reporting intraclass correlation coefficients for reliability
  research}.
\newblock \emph{Journal of chiropractic medicine}, 15(2).

\bibitem[{Kumar et~al.(2022)Kumar, Shrotriya, Sahu, Mishra, Dabre, Puduppully,
  Kunchukuttan, Khapra, and Kumar}]{kumar-etal-2022-indicnlg}
Aman Kumar, Himani Shrotriya, Prachi Sahu, Amogh Mishra, Raj Dabre, Ratish
  Puduppully, Anoop Kunchukuttan, Mitesh~M. Khapra, and Pratyush Kumar. 2022.
\newblock \href {https://aclanthology.org/2022.emnlp-main.360} {{I}ndic{NLG}
  benchmark: Multilingual datasets for diverse {NLG} tasks in {I}ndic
  languages}.
\newblock In \emph{Proceedings of the 2022 Conference on Empirical Methods in
  Natural Language Processing}.

\bibitem[{Kunchukuttan(2020)}]{kunchukuttan2020indicnlp}
Anoop Kunchukuttan. 2020.
\newblock \href {https://github.com/anoopkunchukuttan/indic_nlp_library} {{The
  IndicNLP Library}}.
\newblock {github.com}.

\bibitem[{Ladhak et~al.(2020)Ladhak, Durmus, Cardie, and
  McKeown}]{ladhak-etal-2020-wikilingua}
Faisal Ladhak, Esin Durmus, Claire Cardie, and Kathleen McKeown. 2020.
\newblock \href {https://doi.org/10.18653/v1/2020.findings-emnlp.360}
  {{W}iki{L}ingua: A new benchmark dataset for cross-lingual abstractive
  summarization}.
\newblock In \emph{Findings of the Association for Computational Linguistics:
  EMNLP 2020}.

\bibitem[{Lin(2004)}]{lin-2004-rouge}
Chin-Yew Lin. 2004.
\newblock \href {https://aclanthology.org/W04-1013} {{ROUGE}: A package for
  automatic evaluation of summaries}.
\newblock In \emph{Text Summarization Branches Out}.

\bibitem[{Nallapati et~al.(2016)Nallapati, Zhou, dos Santos,
  {G\"{u}l\c{c}ehre}, and Xiang}]{nallapati-etal-2016-abstractive}
Ramesh Nallapati, Bowen Zhou, Cicero dos Santos, {\c{C}}a{\u{g}}lar
  {G\"{u}l\c{c}ehre}, and Bing Xiang. 2016.
\newblock \href {https://doi.org/10.18653/v1/K16-1028} {Abstractive text
  summarization using sequence-to-sequence {RNN}s and beyond}.
\newblock In \emph{Proceedings of the 20th {SIGNLL} Conference on Computational
  Natural Language Learning}.

\bibitem[{Napoles et~al.(2012)Napoles, Gormley, and
  Van~Durme}]{napoles-etal-2012-annotated2}
Courtney Napoles, Matthew Gormley, and Benjamin Van~Durme. 2012.
\newblock \href {https://aclanthology.org/W12-3018} {Annotated {G}igaword}.
\newblock In \emph{Proceedings of the Joint Workshop on Automatic Knowledge
  Base Construction and Web-scale Knowledge Extraction}.

\bibitem[{Narayan et~al.(2018)Narayan, Cohen, and
  Lapata}]{narayan-etal-2018-dont}
Shashi Narayan, Shay~B. Cohen, and Mirella Lapata. 2018.
\newblock \href {https://doi.org/10.18653/v1/D18-1206} {Don{'}t give me the
  details, just the summary! topic-aware convolutional neural networks for
  extreme summarization}.
\newblock In \emph{Proceedings of the 2018 Conference on Empirical Methods in
  Natural Language Processing}.

\bibitem[{Papineni et~al.(2002)Papineni, Roukos, Ward, and
  Zhu}]{papineni-etal-2002-bleu}
Kishore Papineni, Salim Roukos, Todd Ward, and Wei-Jing Zhu. 2002.
\newblock \href {https://doi.org/10.3115/1073083.1073135} {{B}leu: a method for
  automatic evaluation of machine translation}.
\newblock In \emph{Proceedings of the 40th Annual Meeting of the Association
  for Computational Linguistics}.

\bibitem[{Post(2018)}]{post-2018-call}
Matt Post. 2018.
\newblock \href {https://doi.org/10.18653/v1/W18-6319} {A call for clarity in
  reporting {BLEU} scores}.
\newblock In \emph{Proceedings of the Third Conference on Machine Translation:
  Research Papers}.

\bibitem[{Rush et~al.(2015)Rush, Chopra, and Weston}]{rush-etal-2015-neural}
Alexander~M. Rush, Sumit Chopra, and Jason Weston. 2015.
\newblock \href {https://doi.org/10.18653/v1/D15-1044} {A neural attention
  model for abstractive sentence summarization}.
\newblock In \emph{Proceedings of the 2015 Conference on Empirical Methods in
  Natural Language Processing}.

\bibitem[{Satapara et~al.(2022)Satapara, Modha, Modha, and
  Mehta}]{satapara2022fire}
Shrey Satapara, Bhavan Modha, Sandip Modha, and Parth Mehta. 2022.
\newblock \href {https://dl.acm.org/doi/abs/10.1145/3574318.3574328} {{FIRE}
  2022 {ILSUM} track: Indian language summarization}.
\newblock In \emph{Proceedings of the 14th Annual Meeting of the Forum for
  Information Retrieval Evaluation}.

\bibitem[{Scialom et~al.(2020)Scialom, Dray, Lamprier, Piwowarski, and
  Staiano}]{scialom-etal-2020-mlsum2}
Thomas Scialom, Paul-Alexis Dray, Sylvain Lamprier, Benjamin Piwowarski, and
  Jacopo Staiano. 2020.
\newblock \href {https://doi.org/10.18653/v1/2020.emnlp-main.647} {{MLSUM}: The
  multilingual summarization corpus}.
\newblock In \emph{Proceedings of the 2020 Conference on Empirical Methods in
  Natural Language Processing}.

\bibitem[{Shrout and Fleiss(1979)}]{shrout1979intraclass}
Patrick~E Shrout and Joseph~L Fleiss. 1979.
\newblock \href {https://doi.org/10.1037/0033-2909.86.2.420} {Intraclass
  correlations: uses in assessing rater reliability.}
\newblock \emph{Psychological bulletin}, 86(2).

\bibitem[{Tang et~al.(2021)Tang, Tran, Li, Chen, Goyal, Chaudhary, Gu, and
  Fan}]{tang-etal-2021-multilingual}
Yuqing Tang, Chau Tran, Xian Li, Peng-Jen Chen, Naman Goyal, Vishrav Chaudhary,
  Jiatao Gu, and Angela Fan. 2021.
\newblock \href {https://doi.org/10.18653/v1/2021.findings-acl.304}
  {Multilingual translation from denoising pre-training}.
\newblock In \emph{Findings of the Association for Computational Linguistics:
  ACL-IJCNLP 2021}.

\bibitem[{Taori et~al.(2023)Taori, Gulrajani, Zhang, Dubois, Li, Guestrin,
  Liang, and Hashimoto}]{alpaca2023}
Rohan Taori, Ishaan Gulrajani, Tianyi Zhang, Yann Dubois, Xuechen Li, Carlos
  Guestrin, Percy Liang, and Tatsunori~B. Hashimoto. 2023.
\newblock \href {https://github.com/tatsu-lab/stanford_alpaca} {Stanford
  {A}lpaca: An instruction-following {LLaMA} model}.
\newblock {github.com}.

\bibitem[{Taunk and Varma(2022)}]{taunk2022summarizing}
Dhaval Taunk and Vasudeva Varma. 2022.
\newblock \href {https://arxiv.org/pdf/2303.16657v1.pdf} {Summarizing indian
  languages using multilingual transformers based models}.
\newblock In \emph{Working Notes of FIRE 2022 - Forum for Information Retrieval
  Evaluation}.

\bibitem[{Touvron et~al.(2023)Touvron, Lavril, Izacard, Martinet, Lachaux,
  Lacroix, Rozi{\`e}re, Goyal, Hambro, Azhar et~al.}]{touvron2023llama}
Hugo Touvron, Thibaut Lavril, Gautier Izacard, Xavier Martinet, Marie-Anne
  Lachaux, Timoth{\'e}e Lacroix, Baptiste Rozi{\`e}re, Naman Goyal, Eric
  Hambro, Faisal Azhar, et~al. 2023.
\newblock \href {https://arxiv.org/abs/2302.13971v1} {{LLaMA}: Open and
  efficient foundation language models}.
\newblock \emph{arXiv preprint}.

\bibitem[{Urlana et~al.(2022{\natexlab{a}})Urlana, Bhatt, Surange, and
  Shrivastava}]{urlana2022indian}
Ashok Urlana, Sahil~Manoj Bhatt, Nirmal Surange, and Manish Shrivastava.
  2022{\natexlab{a}}.
\newblock \href {https://arxiv.org/pdf/2303.14461v1.pdf} {Indian language
  summarization using pretrained sequence-to-sequence models}.
\newblock In \emph{Working Notes of FIRE 2022 - Forum for Information Retrieval
  Evaluation}.

\bibitem[{Urlana et~al.(2022{\natexlab{b}})Urlana, Surange, Baswani, Ravva, and
  Shrivastava}]{urlana-etal-2022-tesum}
Ashok Urlana, Nirmal Surange, Pavan Baswani, Priyanka Ravva, and Manish
  Shrivastava. 2022{\natexlab{b}}.
\newblock \href {https://aclanthology.org/2022.lrec-1.614} {{T}e{S}um:
  Human-generated abstractive summarization corpus for {T}elugu}.
\newblock In \emph{Proceedings of the Thirteenth Language Resources and
  Evaluation Conference}.

\bibitem[{Varab and Schluter(2021)}]{varab-schluter-2021-massivesumm}
Daniel Varab and Natalie Schluter. 2021.
\newblock \href {https://doi.org/10.18653/v1/2021.emnlp-main.797}
  {{M}assive{S}umm: a very large-scale, very multilingual, news summarisation
  dataset}.
\newblock In \emph{Proceedings of the 2021 Conference on Empirical Methods in
  Natural Language Processing}.

\bibitem[{Verma et~al.(2023)Verma, Jangra, Verma, and
  Saha}]{verma-etal-2023-large}
Yash Verma, Anubhav Jangra, Raghvendra Verma, and Sriparna Saha. 2023.
\newblock \href {https://aclanthology.org/2023.eacl-main.263} {Large scale
  multi-lingual multi-modal summarization dataset}.
\newblock In \emph{Proceedings of the 17th Conference of the European Chapter
  of the Association for Computational Linguistics}.

\bibitem[{Wang et~al.(2022)Wang, Meng, Zheng, Liang, Li, Qu, and
  Zhou}]{tacl-cross-lingual-survey}
Jiaan Wang, Fandong Meng, Duo Zheng, Yunlong Liang, Zhixu Li, Jianfeng Qu, and
  Jie Zhou. 2022.
\newblock \href {https://doi.org/10.1162/tacl_a_00520} {{A Survey on
  Cross-Lingual Summarization}}.
\newblock \emph{Transactions of the Association for Computational Linguistics},
  10:1304--1323.

\bibitem[{Zhang et~al.(2023)Zhang, Ladhak, Durmus, Liang, McKeown, and
  Hashimoto}]{zhang2023benchmarking}
Tianyi Zhang, Faisal Ladhak, Esin Durmus, Percy Liang, Kathleen McKeown, and
  Tatsunori~B Hashimoto. 2023.
\newblock \href {https://arxiv.org/abs/2301.13848v1} {Benchmarking large
  language models for news summarization}.
\newblock \emph{arXiv preprint}.

\bibitem[{Zhao and Chen(2022)}]{zhao-chen-2022-adapt}
Zheng Zhao and Pinzhen Chen. 2022.
\newblock \href {https://aclanthology.org/2022.ccl-1.73} {To adapt or to
  fine-tune: A case study on abstractive summarization}.
\newblock In \emph{Proceedings of the 21st Chinese National Conference on
  Computational Linguistics}.

\bibitem[{Zhao et~al.(2020)Zhao, Cohen, and Webber}]{zhao-etal-2020-reducing}
Zheng Zhao, Shay~B. Cohen, and Bonnie Webber. 2020.
\newblock \href {https://doi.org/10.18653/v1/2020.findings-emnlp.203} {Reducing
  quantity hallucinations in abstractive summarization}.
\newblock In \emph{Findings of the Association for Computational Linguistics:
  EMNLP 2020}.

\bibitem[{Zhu et~al.(2019)Zhu, Wang, Wang, Zhou, Zhang, Wang, and
  Zong}]{zhu-etal-2019-ncls2}
Junnan Zhu, Qian Wang, Yining Wang, Yu~Zhou, Jiajun Zhang, Shaonan Wang, and
  Chengqing Zong. 2019.
\newblock \href {https://doi.org/10.18653/v1/D19-1302} {{NCLS}: Neural
  cross-lingual summarization}.
\newblock In \emph{Proceedings of the 2019 Conference on Empirical Methods in
  Natural Language Processing and the 9th International Joint Conference on
  Natural Language Processing}.

\end{thebibliography}
\bibliographystyle{acl_natbib}

\newpage
\appendix

\section{Data filtering details}\label{sec:filtering_numbers}
We manually inspected the data and discovered that articles labeled as English may contain non-English content, despite having English URLs. We used these URLs as references to match cross-lingual pairs, but we didn't assume the content was always in English. Therefore, we cleaned our data strictly by language. Table~\ref{tab:filtration_counts} shows the number of invalid samples removed at each step, together with the raw size and the cleaned size.

\section{Data size breakdown}\label{appendix:data_counts}

Table~\ref{tab:cls_counts} lists the total size for each language pair. Rows correspond to the document language and columns correspond to the summary language. Validation and test sizes are 116 each, and the training split employs the rest.

\section{LaBSE similarity breakdown}\label{appendix:labse_scores}

Technically, we obtain LaBSE representations by feeding a summary or document into the model, truncated at the maximum length if necessary. Table~\ref{tab:labse_scores} presents the LaBSE scores between cross-lingual documents and between cross-lingual summaries. Rows and columns indicate the language pair. The upper right triangle contains scores for summary pairs, and the lower left triangle contains scores for document pairs. Due to the lack of support for Manipuri in multilingual BERT \cite{devlin-etal-2019-bert2}, scores involving Manipuri are considerably lower than other language combinations.

\section{Human evaluation details and samples}\label{appendix:data_example}

We carry out a human evaluation to compare headlines and first sentences as summaries. For each of English, Hindi, and Telugu, we ask three native speakers to consider the accuracy, informativeness, and quality of the potential summaries from 50 samples based on the following guideline.

\begin{table}[ht]
    \centering\small
    \begin{tabular}{lccc}
\toprule
& \texttt{\textbf{en}} & \texttt{\textbf{hi}} & \texttt{\textbf{te}} \\
\midrule
    Headlines       & 0.80        & 0.52        & 0.54        \\
    First sentences & 0.82        & 0.88        & 0.67        \\
\bottomrule
    \end{tabular}
    \caption{Inter rater reliability (ICC).}\label{tab:irr}
\end{table}

\noindent\framebox{%
\parbox{0.475\textwidth}{
In this task, you will look at a set of articles and summaries. Each article will be presented with two (2) potential summaries. The aim of the evaluation is to identify whether the presented text is a summary. Below are the steps you need to take.
\begin{enumerate}
    \item Read the article to understand the context.
    \item Read both summaries carefully. Consider the accuracy, informativeness, and quality of the summaries. We provide the definition of these criteria below. \begin{itemize}\setlength\itemsep{0ex}\setlength{\itemindent}{-2.5ex}
        \item \textit{Accuracy}: how closely the summary reflects the factual content of a news article. An accurate summary should not include any false information or misrepresentations of the content.
        \item \textit{Informativeness}: how much information the summary provides about the main points of the news article. An informative summary should provide enough detail to give the reader a good understanding of the article's content.
        \item \textit{Quality}: the overall standard of the summary. A high-quality summary should be fluent, well-written, free of errors, and easy to read.
    \end{itemize}\setlength\itemsep{0ex}
    \item Provide your binary decision on whether each text is a summary. Base your decision solely on the quality, accuracy, and informativeness of the content, without being influenced by factors such as writing style, personal preferences, etc. Please note that an incomplete sentence is acceptable as long as it does not compromise accuracy, informativeness, or quality.
\end{enumerate}
}%
}
\\

For each sample document, we record the majority voting (yes/no) from three annotators in Section~\ref{sec:summary_quality} Table~\ref{tab:human_eval_yes_no}. Alongside the results, in Table~\ref{tab:irr}, we compute the inter-rater reliability using Intra-class Correlation Coefficient as per \citep[ICC,][]{koo2016guideline,shrout1979intraclass}. We observe that the annotators are in moderate to good agreement for all languages and both potential summaries.

We also note samples from the dataset to show potential problems with using first sentences as summaries in Table~\ref{fig:example1}. Sometimes the first sentence is part of a speech expressing gratitude.

\begin{table*}[thb]
\centering\small
\setlength{\tabcolsep}{0.95ex}
\begin{tabular}{lrrrrrrrrrrrrrr}
\toprule
& \multicolumn{1}{c}{\textbf{\texttt{as}}}
& \multicolumn{1}{c}{\textbf{\texttt{bn}}}
& \multicolumn{1}{c}{\textbf{\texttt{gu}}}
& \multicolumn{1}{c}{\textbf{\texttt{hi}}}
& \multicolumn{1}{c}{\textbf{\texttt{kn}}}
& \multicolumn{1}{c}{\textbf{\texttt{ml}}}
& \multicolumn{1}{c}{\textbf{\texttt{mni}}}
& \multicolumn{1}{c}{\textbf{\texttt{mr}}}
& \multicolumn{1}{c}{\textbf{\texttt{or}}}
& \multicolumn{1}{c}{\textbf{\texttt{pa}}}
& \multicolumn{1}{c}{\textbf{\texttt{ta}}}
& \multicolumn{1}{c}{\textbf{\texttt{te}}}
& \multicolumn{1}{c}{\textbf{\texttt{ur}}}
& \multicolumn{1}{c}{\textbf{\texttt{en}}} \\
\midrule
raw size & 2925 & 9250 & 7393 & 7642 & 5743 & 5106 & 5432 & 6034 & 4760 & 7315 & 6314 & 6656 & 5622 & 13844 \\
 - language mismatch & 68 & 3418 & 302 & 80 & 361 & 344 & 271 & 93 & 40 & 2367 & 68 & 422 & 1144 & 5288 \\
 - duplicate pairs & 4 & 11 & 11 & 5 & 16 & 7 & 1 & 12 & 8 & 11 & 23 & 8 & 6 & 2 \\
 - duplicate summaries & 8 & 28 & 34 & 110 & 38 & 25 & 14 & 54 & 26 & 53 & 54 & 34 & 24 & 202 \\
 - empty & 0 & 1 & 0 & 0 & 0 & 0 & 2 & 0 & 1 & 0 & 0 & 0 & 0 & 0 \\
 - prefixes & 33 & 37 & 28 & 49 & 13 & 52 & 92 & 28 & 13 & 30 & 73 & 46 & 133 & 169 \\
 - doc \textless  2 sentences & 723 & 191 & 216 & 35 & 5 & 12 & 113 & 30 & 21 & 40 & 13 & 19 & 0 & 19 \\
 - sum \textless  3 tokens & 0 & 7 & 0 & 0 & 3 & 1 & 3 & 1 & 0 & 0 & 4 & 1 & 0 & 4 \\
\midrule
\textbf{final (monolingual)} & 2089 & 5557 & 6802 & 7363 & 5307 & 4665 & 4936 & 5816 & 4651 & 4814 & 6079 & 6126 & 4315 & 8160 \\
\bottomrule
\end{tabular}
\caption{Data preprocessing and filtering statistics for document-summary pairs in each language.}
\label{tab:filtration_counts}
\end{table*}
 
\begin{table*}[ht]
\centering\small
\begin{tabular}{rcccccccccccccc}
\toprule
 \diagbox[width=8ex]{$D_{Li}$}{$S_{Lj}$} & \textbf{\texttt{as}} & \textbf{\texttt{bn}} & \textbf{\texttt{gu}} & \textbf{\texttt{hi}} & \textbf{\texttt{kn}} & \textbf{\texttt{ml}} & \textbf{\texttt{mni}} & \textbf{\texttt{mr}} & \textbf{\texttt{or}} & \textbf{\texttt{pa}} & \textbf{\texttt{ta}} & \textbf{\texttt{te}} & \textbf{\texttt{ur}} & \textbf{\texttt{en}} \\
\midrule
\textbf{\texttt{as}} & 2089  & 1153 & 1728 & 1716 & 1411 & 1166 & 1737 & 1415 & 1711 & 1332 & 1607 & 1570 & 1571 & 1743 \\ 
\textbf{\texttt{bn}} & 1153 & 5557  & 4655 & 3987 & 3573 & 3332 & 2898 & 4268 & 3079 & 3479 & 3845 & 4239 & 2345 & 4075 \\ 
\textbf{\texttt{gu}} & 1728 & 4655 & 6802 & 4896 & 4411 & 3901 & 3998 & 4940 & 4021 & 4052 & 4731 & 5111 & 3185 & 5216 \\ 
\textbf{\texttt{hi}} & 1716 & 3987 & 4896 & 7363   & 3680 & 3723 & 3163 & 4340 & 3784 & 3426 & 4817 & 4376 & 3061 & 5899 \\ 
\textbf{\texttt{kn}} & 1411 & 3573 & 4411 & 3680 & 5307  & 3176 & 3405 & 3800 & 3439 & 3526 & 3720 & 4151 & 2742 & 4325 \\ 
\textbf{\texttt{ml}} & 1166 & 3332 & 3901 & 3723 & 3176 & 4665  & 2423 & 3547 & 3037 & 2881 & 3957 & 3737 & 2299 & 3769 \\ 
\textbf{\texttt{mni}} & 1737 & 2898 & 3998 & 3163 & 3405 & 2423 & 4936   & 3120 & 2901 & 3198 & 3181 & 3596 & 3199 & 3807 \\ 
\textbf{\texttt{mr}} & 1415 & 4268 & 4940 & 4340 & 3800 & 3547 & 3120 & 5816  & 3450 & 3573 & 4219 & 4488 & 2582 & 4411 \\ 
\textbf{\texttt{or}} & 1711 & 3079 & 4021 & 3784 & 3439 & 3037 & 2901 & 3450 & 4651 & 3224 & 3722 & 3667 & 2646 & 3873 \\ 
\textbf{\texttt{pa}} & 1332 & 3479 & 4052 & 3426 & 3526 & 2881 & 3198 & 3573 & 3224 & 4814  & 3356 & 3772 & 2489 & 3615 \\ 
\textbf{\texttt{ta}} & 1607 & 3845 & 4731 & 4817 & 3720 & 3957 & 3181 & 4219 & 3722 & 3356 & 6079   & 4514 & 3139 & 4909 \\ 
\textbf{\texttt{te}} & 1570 & 4239 & 5111 & 4376 & 4151 & 3737 & 3596 & 4488 & 3667 & 3772 & 4514 & 6126  & 2847 & 4955 \\ 
\textbf{\texttt{ur}} & 1571 & 2345 & 3185 & 3061 & 2742 & 2299 & 3199 & 2582 & 2646 & 2489 & 3139 & 2847 & 4315  & 3415 \\ 
\textbf{\texttt{en}} & 1743 & 4075 & 5216 & 5899 & 4325 & 3769 & 3807 & 4411 & 3873 & 3615 & 4909 & 4955 & 3415 & 8160   \\ 
\bottomrule
\end{tabular}
\caption{PMIndiaSum data sizes for all language pairs. Each cell corresponds to the size of a 
dataset $({D}_{Li},{S}_{Lj})$, with source document $D$ in language $Li$ and target summary $S$ in language $Lj$.}
\label{tab:cls_counts}
\end{table*}

\begin{table*}[ht]
\centering\small
\begin{tabular}{rcccccccccccccc} 
\toprule
 \diagbox[width=8ex]{$Li$}{$Lj$}  & \textbf{\texttt{as}} & \textbf{\texttt{bn}} & \textbf{\texttt{gu}} & \textbf{\texttt{hi}} & \textbf{\texttt{kn}} & \textbf{\texttt{ml}} & \textbf{\texttt{mni}} & \textbf{\texttt{mr}} & \textbf{\texttt{or}} & \textbf{\texttt{pa}} & \textbf{\texttt{ta}} & \textbf{\texttt{te}} & \textbf{\texttt{ur}} & \textbf{\texttt{en}} \\
\midrule
\textbf{\texttt{as}} & - &  0.88  & 0.86 & 0.82 & 0.87 & 0.87 & 0.80 & 0.85 & 0.90 & 0.87 & 0.85 & 0.88 & 0.85 & 0.87 \\
\textbf{\texttt{bn}} & 0.89 & - & 0.86 & 0.85 & 0.87 & 0.86 & 0.73 & 0.87 & 0.88 & 0.87 & 0.85 & 0.86 & 0.87 & 0.86 \\
\textbf{\texttt{gu}} & 0.85 & 0.90 & - & 0.92 & 0.91 & 0.88 & 0.70 & 0.90 & 0.90 & 0.92 & 0.88 & 0.90 & 0. 91 & 0.89 \\
\textbf{\texttt{hi}} & 0.81 & 0.89  & 0.93 & - & 0.89 & 0.86 & 0.65 & 0.91 & 0.87 & 0.91 & 0.88 & 0.88 & 0.90 & 0.89 \\
\textbf{\texttt{kn}} & 0.86 & 0.91 & 0.94 & 0.92 & - & 0.89 & 0.72 & 0.90 & 0.89 & 0.91 & 0.89 & 0.91 & 0.89 & 0.90   \\
\textbf{\texttt{ml}} & 0.88 & 0.90 & 0.91 & 0.89 & 0.93& - & 0.73 & 0.88 & 0.89 & 0.89 & 0.87 & 0.88 & 0.88 & 0.88 \\
\textbf{\texttt{mni}} & 0.84 & 0.75 & 0.71 & 0.68 & 0.73 & 0.75 & - & 0.69 & 0.74 & 0.71 & 0.71 & 0.74 & 0.69 & 0.74 \\
\textbf{\texttt{mr}} & 0.85 & 0.90 & 0.93 & 0.93 & 0.92 & 0.90 & 0.71 & - & 0.89 & 0.90 & 0.88 & 0.89 & 0.89 & 0.89\\
\textbf{\texttt{or}} & 0.90 & 0.91 & 0.92 & 0.90 & 0.92 & 0.92 & 0.76 & 0.91 & - & 0.90 & 0.88 & 0.89 & 0.88 & 0.90 \\
\textbf{\texttt{pa}} & 0.87 & 0.90 & 0.94 & 0.93 & 0.93 & 0.91 & 0.73 & 0.92 & 0.92 & - & 0.89 & 0.90  & 0.91  & 0.90 \\
\textbf{\texttt{ta}} & 0.85 & 0.90 & 0.91 & 0.90 & 0.93 & 0.91 & 0.73 & 0.91 & 0.91 & 0.91 & - & 0.89  & 0.88  & 0.88 \\
\textbf{\texttt{te}} & 0.87 &  0.91 & 0.92 & 0.90 & 0.94 & 0.92 & 0.75 & 0.91 & 0.92 & 0.93 & 0.92 & -  & 0.89  & 0.90 \\
\textbf{\texttt{ur}} & 0.84 & 0.89 & 0.92 & 0.90 & 0.91 & 0.90 & 0.72 & 0.90 & 0.90 & 0.92 & 0.90 & 0.91  & -  & 0.90 \\
\textbf{\texttt{en}} & 0.84 & 0.89 & 0.90 & 0.91 & 0.91 & 0.90 & 0.72 & 0.90 & 0.90 & 0.90 & 0.90 & 0.90  & 0.89  & - \\
\bottomrule
\end{tabular}
\caption{LaBSE scores between parallel documents ($D_{Li}, D_{Lj}$, bottom left) and parallel summaries ($S_{Li}, S_{Lj}$, upper right).}
\label{tab:labse_scores}
\end{table*}

\section{Experimental setup}\label{appendix:exp_setup}

For prompting, our initial finding is that open-source LLMs do not work very well for non-English, so our prompting experiments are for English only. In terms of technical details, IndicBART and mBART-large-50 models have 244M and 610M parameters each. For fine-tuning, we set a training budget of 100 epochs and apply early stopping after three consecutive non-improving cross-entropy during validation. We use an effective batch size of 96 by combining different batch sizes, numbers of GPUs, and gradient accumulation. For document-summary pairs, we set the maximum lengths to 1024 and 64 respectively. All other configurations follow the Hugging Face trainer default.\footnote{\href{https://huggingface.co/docs/transformers/main\_classes/trainer}{https://huggingface.co/docs/transformers/main\_classes/\\trainer}} We do not perform hyperparameter search. Specifically, our GPUs include Nvidia GeForce RTX 2080 Ti (11GB), GeForce RTX 3090 (24GB), and A100 (80GB). Models mostly converge within 10 epochs.

\section{Other summarization techniques}\label{appendix:adapter_other_model}

\subsection{Adapters}
Adapters are small trainable modules inserted into a giant PLM; during PLM fine-tuning, the entire model can be frozen except for the adapters \citep{pmlr-v97-houlsby19a}. This paradigm achieves great parameter efficiency as only the adapter weights need to be updated and saved.

Adapters are relevant to summarization research, but we omit experiments on these because \citet{zhao-chen-2022-adapt} show that with a few thousand training data, the performance of adapters is not comparable to fine-tuning a PLM. We, nevertheless, encourage future research to try this out.

\subsection{Models trained on another dataset}
An effortless approach apart from extracting the first sentence from a document is to use a summarization model that has been trained on another dataset. In this regard, we experiment with an mT5-based multilingual summarization model trained on the XL-Sum dataset comprising 44 languages.\footnote{\href{https://huggingface.co/csebuetnlp/mT5\_multilingual\_XLSum}{https://huggingface.co/csebuetnlp/mT5\_multilingual/\\\_XLSum}} We tested it on our PMIndiaSum monolingual test sets. As Table~\ref{tab:xlsum_trained} shows, the performance of the XL-Sum pre-trained model is subpar on our data, potentially due to domain mismatch. This implies that our test sets cannot be easily solved by training on other datasets.

\begin{table}[ht]
\centering\small
\begin{tabular}{cccc}
\toprule
& {R2} & {RL} & {BL} \\
\midrule
\texttt{\textbf{bn}} & 16.9         & 30.0          &\phantom{0}9.8           \\ 
\texttt{\textbf{gu}} & 21.2         & 33.3         & 15.0            \\ 
\texttt{\textbf{hi}} & 23.4         & 40.1         & 16.4          \\ 
\texttt{\textbf{mr}} & 21.1         & 36.2         & 14.8          \\ 
\texttt{\textbf{pa}} & 29.7         & 45.1         & 23.2          \\ 
\texttt{\textbf{ta}} & 19.7         & 35.6         & 12.8          \\ 
\texttt{\textbf{te}} & \phantom{0}7.7         &  17.9         & \phantom{0}4.1           \\ 
\texttt{\textbf{ur}} & 26.3         & 40.4         & 20.1          \\ 
\texttt{\textbf{en}} & 11.2         & 24.0         & \phantom{0}5.5           \\ \bottomrule
\end{tabular}
\caption{Results of a multilingual model trained on XL-Sum and tested on our PMIndiaSum.}
\label{tab:xlsum_trained}
\end{table}

\section{Standard deviations}\label{appendix:std_scores}
Standard deviations are listed in Table~\ref{tab:mono_std_results} for monolingual summarization, Table~\ref{tab:cross_results_std} for cross-lingual summarization, Table~\ref{tab:multilingual-indicbart-std} for multilingual IndicBART, and Table~\ref{tab:multilingual-mbart-std} for multilingual mBART. These correspond to the mean ROUGE and BLEU scores reported in Section~\ref{sec:results}. We do not notice any peculiar numbers.

\begin{table}[ht]
\centering
\small
\begin{tabular}{rcccccccc} 
\toprule
& \multicolumn{3}{c}{\textbf{IndicBART}} & \multicolumn{3}{c}{\textbf{mBART}} \\
\cmidrule(lr){2-4}\cmidrule(lr){5-7}
 & R2 & RL & BL & R2 & RL & BL  \\
\midrule
\textbf{\texttt{as}} & 0.6 & 0.5 & 0.6 & - & - & - \\
\textbf{\texttt{bn}} & 1.0 & 1.0 & 1.0 & 0.2 & 0.3 & 0.7 \\
\textbf{\texttt{gu}} & 1.5 & 1.4 & 0.1 & 1.3 & 1.5 & 1.6 \\
\textbf{\texttt{hi}} & 0.8 & 0.8 & 0.6 & 0.5 & 0.4 & 1.8  \\
\textbf{\texttt{kn}} & 0.6 & 0.8 & 0.2 & - & - & - \\
\textbf{\texttt{ml}} & 0.8 & 0.6 & 0.3 & 1.6 & 1.4 & 1.1 \\
\textbf{\texttt{mni}} & 0.3 & 0.7 & 0.5 & 1.9 & 2.7 & 1.7 \\
\textbf{\texttt{mr}} & 1.4 & 1.4 & 2.7  & 1.3 & 1.1 & 1.2 \\
\textbf{\texttt{or}} & 1.5 & 1.2 & 1.5 & - & - & - \\
\textbf{\texttt{pa}} & 0.7 & 0.6 & 0.7 & - & - & - \\
\textbf{\texttt{ta}} & 0.9 & 0.5 & 0.2 & 0.7 & 0.4 & 2.1 \\
\textbf{\texttt{te}} & 0.4 & 0.9 & 0.0 & 1.3 & 1.1 & 0.7  \\
\textbf{\texttt{ur}} & - & - & - & 2.8 & 2.1 & 2.7 \\
\textbf{\texttt{en}} & 0.4 & 0.00 & 1.8 & 2.0 & 1.1 & 2.1 \\
\bottomrule
\end{tabular}
\caption{Monolingual benchmark: standard deviations.}
\label{tab:mono_std_results}
\end{table}

\begin{table*}[htb]
    \centering\small
    \begin{tabular}{c}
    \toprule
    \includegraphics[width=6in,clip]{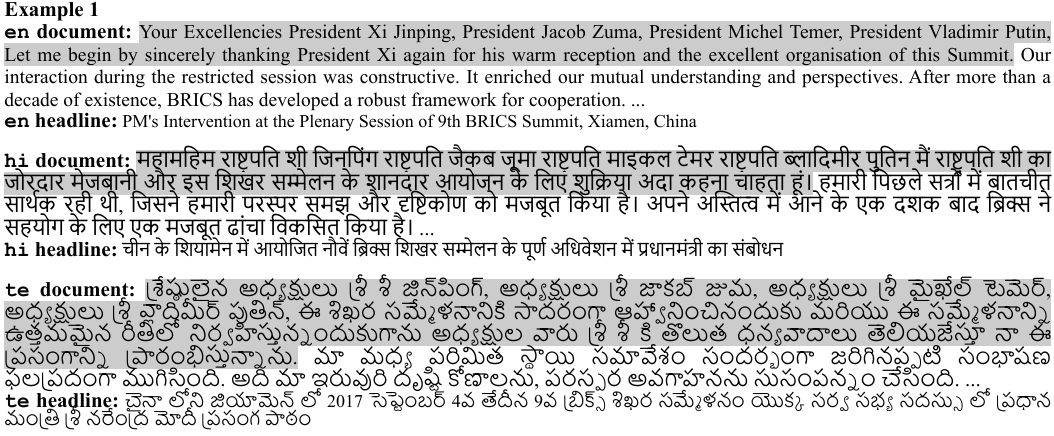} \\
    \midrule
    \includegraphics[width=6in,clip]{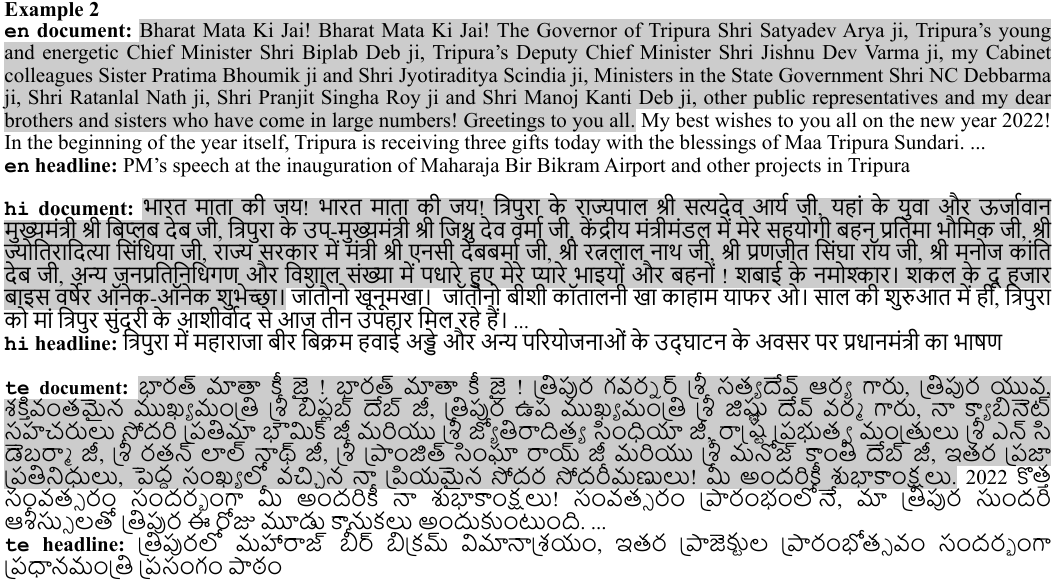} \\
    \bottomrule
    \end{tabular}
    \caption{Two examples of PMIndiaSum document-headline pairs in English, Hindi, and Telugu. The documents' \colorbox{gray!40}{first sentences} could be part of a speech, which poses a problem when used as a summary, whereas the headline is a more appropriate summary.}
    \label{fig:example1}
\end{table*}

\section{Error analysis details}\label{appendix:error_analysis}

We conduct an error analysis on the outputs of monolingual and multilingual models for English, Hindi, and Telugu. Additionally, we analyzed the outputs of cross-lingual and multilingual models for English-to-Hindi and Hindi-to-English summarization. We define our error categories in Table~\ref{tab:error_categories} which displays each error type with an explanation and an example. As per our definition, a single summary output may fall into multiple error categories.

It is important to highlight the distinction between this error analysis and summary quality analysis in previous works \citep{grusky-etal-2018-newsroom2, clark2023seahorse}. In our analysis, human annotators identify errors in the system outputs by referring to the gold output rather than the source document. To ensure accuracy, our annotators are native speakers of the language of the summaries. In the case of cross-lingual summaries, they were also bilingual in the source language.

\begin{table*}[htb]
\centering\small
\begin{tabularx}{0.97\textwidth}{lrlX}
\toprule
\multicolumn{1}{l}{\hspace*{2ex}\textbf{Error type}} & \multicolumn{3}{l}{\hspace*{16ex}\textbf{Error description}}   \\
\midrule
\multirow{1}{*}{\makecell{Comprehensibility}} & \multicolumn{3}{l}{\textbf{The output cannot be understood: wrong language, empty string, etc.}} \\
& E.g. & {Reference:} & Cabinet approves creation of one post each of Vice-Chairperson and Member in the National \\
& & {Output:} & Cabinet approves creation of two-on-one posts of the National Cleaning Employees Commission \\
\midrule
\multirow{1}{*}{\makecell{Grammar \& fluency}} & \multicolumn{3}{l}{\textbf{The output has grammatical errors or is unnatural and awkward.}}     \\
& E.g. & {Reference:} & PM to dedicate naval submarine INS Kalvari to the nation tomorrow \\
& & {Output:} & PM to dedicate Navy Navy Navy Navy Navy Navy Navy Navy Navy Navy \\
\midrule
\multirow{1}{*}{{Factuality}}       & \multicolumn{3}{l}{\textbf{The output contains incorrect details as per the reference: named entity, number, etc.}} \\
& E.g. & {Reference:} & Cabinet apprised of two Bilateral MoUs between India and Cuba, and India and Korea in the area of Biotechnology \\
& & {Output:} & Cabinet apprised of the MoU between India and Mongolia on cooperation in the field of Biotechnology \\
\midrule
\multirow{1}{*}{{Omission}}         & \multicolumn{3}{l}{\textbf{The output does not cover some information in the reference.}}          \\
& E.g. & {Reference:} & PM to hold ‘Samvad’ with Beneficiaries of Pradhan Mantri Bhartiya Janaushadhi Pariyojna and affordable cardiac stents and knee implants on June 7 \\
& & {Output:} & PM to interact with beneficiaries of Pradhan Mantri Bhartiya Janaushadhi Pariyojna on 7th June \\
\midrule
\multirow{1}{*}{{Redundancy}}       & \multicolumn{3}{l}{\textbf{The output consists of repeated or too much information compared to the reference.}}   \\
& E.g. & {Reference:} &  PM to confer Awards for Excellence in Public Administration and address Civil Servants tomorrow \\
& & {Output:} &  PM to confer Awards for Excellence in Public Administration for effective implementation of identified Priority Programs and Innovation to districts/implementing units and other Central/State organisations at Vigyan Bhawan tomorrow \\
\midrule
\multirow{1}{*}{{No error}}    & \multicolumn{3}{l}{\textbf{The output conveys the same information as the reference.}}      \\ 
\bottomrule                               
\end{tabularx}
\caption{Error categories defined in our error analysis.}
\label{tab:error_categories}
\end{table*}

\clearpage
\begin{sidewaystable}[htb]
\vspace{49ex}
\centering\scriptsize
\setlength{\tabcolsep}{1ex}
\begin{tabular}{rrrrrrrrrrrrrrrrrrrrrrrrr}
\toprule
& \multicolumn{6}{c}{\textbf{Summarization-Translation}} & \multicolumn{6}{c}{\textbf{Translation-Summarization}} & \multicolumn{6}{c}{\textbf{Fine-tuning}}  & \multicolumn{6}{c}{\textbf{Zero-shot}}\\
\cmidrule(lr){2-7}\cmidrule(lr){8-13}\cmidrule(lr){14-19}\cmidrule(lr){20-25}
& \multicolumn{3}{c}{IndicBART} & \multicolumn{3}{c}{mBART} & \multicolumn{3}{c}{IndicBART} & \multicolumn{3}{c}{mBART} & \multicolumn{3}{c}{IndicBART} & \multicolumn{3}{c}{mBART} & \multicolumn{3}{c}{IndicBART} & \multicolumn{3}{c}{mBART} \\
\cmidrule(lr){2-4}\cmidrule(lr){5-7}\cmidrule(lr){8-10}\cmidrule(lr){11-13}\cmidrule(lr){14-16}\cmidrule(lr){17-19}\cmidrule(lr){20-22}\cmidrule(lr){23-25}
 & \multicolumn{1}{c}{R2} & \multicolumn{1}{c}{RL} & \multicolumn{1}{c}{BL} & \multicolumn{1}{c}{R2} & \multicolumn{1}{c}{RL} & \multicolumn{1}{c}{BL} & \multicolumn{1}{c}{R2} & \multicolumn{1}{c}{RL} & \multicolumn{1}{c}{BL} & \multicolumn{1}{c}{R2} & \multicolumn{1}{c}{RL} & \multicolumn{1}{c}{BL} & \multicolumn{1}{c}{R2} & \multicolumn{1}{c}{RL} & \multicolumn{1}{c}{BL} & \multicolumn{1}{c}{R2} & \multicolumn{1}{c}{RL} & \multicolumn{1}{c}{BL} & \multicolumn{1}{c}{R2} & \multicolumn{1}{c}{RL} & \multicolumn{1}{c}{BL } & \multicolumn{1}{c}{R2} & \multicolumn{1}{c}{RL} & \multicolumn{1}{c}{BL}\\
 \midrule
\textbf{\texttt{hi-en}}  & 0.4 & 0.5 & 0.4 & 0.5 & 0.4 & 0.6 & 0.5 & 0.6 & 0.7 & 0.9 & 1.1 & 1.2 & 0.3 & 0.2 & 0.7 & 6.9 & 6.0 & 5.9 & 0.0 & 0.1 & 0.0 & 0.4 & 4.3 & 0.1 \\
\textbf{\texttt{en-hi}} & 0.4 & 0.8 & 0.5 & 0.3 & 0.5 & 0.5 & 0.8 & 0.8 & 0.9 & 1.3 & 1.0 & 1.2 & 0.3 & 0.2 & 0.4 & 1.8 & 1.8 & 1.9 & 0.0 & 0.1 & 0.0 & 0.0 & 0.2 & 0.0\\
\textbf{\texttt{gu-te}} & 0.2 & 0.4 & 0.0 & 0.3 & 0.4 & 0.2 & 0.4 & 0.4 & 0.2 & 0.7 & 1.1 & 0.3 & 0.4 & 0.5 & 0.2 & 0.5 & 1.1 & 0.2 & 0.0 & 0.0 & 0.0 & 0.0 & 0.1 & 0.0 \\
\textbf{\texttt{te-gu}} & 0.5 & 0.2 & 0.3 & 0.6 & 1.0 & 0.2 & 0.1 & 0.4 & 0.3 & 0.9 & 1.9 & 0.3 & 0.3 & 0.2 & 0.3 & 0.7 & 0.2 & 0.2 & 0.0 & 0.0 & 0.0 & 0.0 & 0.1 & 0.0 \\
\textbf{\texttt{ml-mni}} & 0.4 & 0.5 & 0.4 & 1.5 & 1.6 & 1.3 & 0.4 & 0.3 & 0.5 & 0.4 & 0.9 & 0.5 & 0.2 & 0.4 & 0.3 & 1.1 & 1.5 & 0.3 & 0.0 & 0.0 & 0.0 & 0.0 & 0.0 & 0.0\\
\textbf{\texttt{mni-ml}} & 0.7 & 0.6 & 0.6 & 1.5 & 2.0 & 0.9 & 0.5 & 0.6 & 0.3 & 0.1 & 0.2 & 0.1 & 0.3 & 0.3 & 0.1 & 0.8 & 3.1 & 0.3 & 0.0 & 0.0 & 0.0 & 0.0 & 0.0 & 0.0 \\
\textbf{\texttt{mr-bn}} & 0.7 & 0.4 & 0.4 & 1.5 & 1.5 & 1.1 & 0.3 & 0.1 & 0.1 & 0.4 & 0.0 & 0.7 & 0.2 & 0.2 & 0.2 & 1.1 & 1.9 & 1.2 & 0.0 & 0.0 & 0.0 & 0.0 & 0.0 & 0.0 \\
\textbf{\texttt{bn-mr}}  &0.2 & 0.2 & 0.3 & 0.8 & 1.3 & 0.4 & 0.7 & 0.8 & 0.5 & 1.1 & 1.8 & 0.6 & 0.2 & 0.1 & 0.0 & 1.7 & 2.4 & 2.3 & 0.0 & 0.0 & 0.0 & 0.0 & 0.0 & 0.0 \\
\textbf{\texttt{te-ta}} & 0.2 & 0.2 & 0.3 & 0.8 & 1.3 & 0.4 & 0.7 & 0.8 & 0.5 & 1.1 & 1.8 & 0.6 & 0.2 & 0.1 & 0.0 & 1.7 & 2.4 & 2.3 & 0.0 & 0.0 & 0.0 & 0.0 & 0.0 & 0.0  \\
\textbf{\texttt{ta-te}}  & 0.5 & 0.6 & 0.5 & 0.6 & 0.4 & 0.4 & 0.9 & 1.0 & 0.4 & 1.2 & 1.1 & 0.3 & 0.4 & 0.6 & 0.4 & 1.3 & 2.2 & 3.3 & 0.0 & 0.0 & 0.0 & 0.0 & 0.0 & 0.0 \\
\textbf{\texttt{mni-en}} &0.5 & 0.6 & 0.3 & 0.2 & 0.6 & 0.3 & 0.1 & 0.1 & 0.1 & 1.5 & 1.2 & 0.9 & 0.5 & 0.5 & 0.4 & 6.0 & 9.4 & 4.3 & 0.0 & 0.0 & 0.0 & 0.4 & 2.7 & 0.0\\
\textbf{\texttt{en-mni}}  & 0.6 & 0.8 & 0.6 & 0.4 & 0.6 & 0.7 & 0.3 & 0.2 & 0.2 & 1.7 & 2.8 & 1.0 & 0.2 & 0.3 & 0.2 & 1.4 & 3.5 & 0.6 & 0.0 & 0.0 & 0.0 & 0.0 & 0.0 & 0.0 \\

\bottomrule
\end{tabular}
\vspace{-1ex}
\caption{Cross-lingual benchmarks: standard deviations.}
\label{tab:cross_results_std}

\vspace{3ex}

\centering\scriptsize
\setlength{\tabcolsep}{1.0ex}
\begin{tabular}{rrrrrrrrrrrrrrrrrrrrrrrrrrrrrrrrrrrrrrrr} 
 \toprule
& \multicolumn{3}{c}{\textbf{\texttt{as}}} & \multicolumn{3}{c}{\textbf{\texttt{bn}}} & \multicolumn{3}{c}{\textbf{\texttt{gu}}} & \multicolumn{3}{c}{\textbf{\texttt{\texttt{hi}}}} & \multicolumn{3}{c}{\textbf{\texttt{kn}}} & \multicolumn{3}{c}{\textbf{\texttt{ml}}} & \multicolumn{3}{c}{\textbf{\texttt{mni}}} & \multicolumn{3}{c}{\textbf{\texttt{mr}}} & \multicolumn{3}{c}{\textbf{\texttt{or}}} & \multicolumn{3}{c}{\textbf{\texttt{pa}}} & \multicolumn{3}{c}{\textbf{\texttt{ta}}} & \multicolumn{3}{c}{\textbf{\texttt{te}}} & \multicolumn{3}{c}{\textbf{\texttt{en}}} \\
\cmidrule(lr){2-4}\cmidrule(lr){5-7}\cmidrule(lr){8-10}\cmidrule(lr){11-13}\cmidrule(lr){14-16}\cmidrule(lr){17-19}\cmidrule(lr){20-22}\cmidrule(lr){23-25}\cmidrule(lr){26-28}\cmidrule(lr){29-31}\cmidrule(lr){32-34}\cmidrule(lr){35-37}\cmidrule(lr){38-40}
 & \multicolumn{1}{c}{R2} & \multicolumn{1}{c}{RL} & \multicolumn{1}{c}{BL} & \multicolumn{1}{c}{R2} & \multicolumn{1}{c}{RL} & \multicolumn{1}{c}{BL} & \multicolumn{1}{c}{R2} & \multicolumn{1}{c}{RL} & \multicolumn{1}{c}{BL} & \multicolumn{1}{c}{R2} & \multicolumn{1}{c}{RL} & \multicolumn{1}{c}{BL} & \multicolumn{1}{c}{R2} & \multicolumn{1}{c}{RL} & \multicolumn{1}{c}{BL} & \multicolumn{1}{c}{R2} & \multicolumn{1}{c}{RL} & \multicolumn{1}{c}{BL} & \multicolumn{1}{c}{R2} & \multicolumn{1}{c}{RL} & \multicolumn{1}{c}{BL} & \multicolumn{1}{c}{R2} & \multicolumn{1}{c}{RL} & \multicolumn{1}{c}{BL} & \multicolumn{1}{c}{R2} & \multicolumn{1}{c}{RL} & \multicolumn{1}{c}{BL} & \multicolumn{1}{c}{R2} & \multicolumn{1}{c}{RL} & \multicolumn{1}{c}{BL} & \multicolumn{1}{c}{R2} & \multicolumn{1}{c}{RL} & \multicolumn{1}{c}{BL} & \multicolumn{1}{c}{R2} & \multicolumn{1}{c}{RL} & \multicolumn{1}{c}{BL} & \multicolumn{1}{c}{R2} & \multicolumn{1}{c}{RL} & \multicolumn{1}{c}{BL} \\
 \midrule
\textbf{\texttt{as}} & 0.8 & 0.9 & 0.4 & 0.5 & 0.5 & 0.4 & 1.3 & 2.1 & 0.8 & 0.3 & 0.3 & 0.1 & 0.3 & 0.5 & 0.2 & 0.3 & 0.2 & 0.0 & 1.1 & 1.6 & 0.7 & 0.2 & 0.5 & 0.1 & 0.7 & 1.4 & 0.3 & 0.6 & 0.7 & 0.3 & 0.4 & 0.6 & 0.2 & 0.7 & 1.1 & 0.2 & 0.7 & 1.1 & 0.3 \\
\textbf{\texttt{bn}} & 0.6 & 1.2 & 0.3 & 0.6 & 0.9 & 0.3 & 0.7 & 1.7 & 0.4 & 0.8 & 1.2 & 0.6 & 0.6 & 1.5 & 0.6 & 0.5 & 1.1 & 0.4 & 0.4 & 0.2 & 0.2 & 0.5 & 0.6 & 0.6 & 0.7 & 0.7 & 0.3 & 1.2 & 1.7 & 0.8 & 1.0 & 1.5 & 0.3 & 0.6 & 0.8 & 0.3 & 0.4 & 0.4 & 0.3\\
\textbf{\texttt{gu}} & 0.1&0.1&0.1&0.3&0.5&0.1&0.4&0.7&0.5&1.3&1.6&1.2&0.7&0.3&0.7&0.1&0.3&0&0.6&1.2&0.5&0.3&0.6&0.4&0.9&1.3&0.3&0.3&0.4&0.3&0.8&0.9&0.2&0.9&1.3&0.3&1.2&2.1&0.7  \\
\textbf{\texttt{hi}} & 0.6&1.1&0.4&0.2&0.6&0.0&0.4&0.9&0.3&0.4&0.7&1.1&1.7&2.3&0.8&0.4&0.3&0.4&0.1&0.4&0.6&1.3&2.0&0.6&1.7&2.5&0.6&1.4&2.3&1.1&0.6&0.9&0.2&0.4&0.6&0.3&0.7&1.2&0.4 \\
\textbf{\texttt{kn}} & 0.3&0.5&0.2&0.3&0.3&0.0&0.9&0.4&0.6&0.9&1.7&0.5&0.8&1.0&0.7&0.8&1.2&0.1&0.6&0.8&0.7&0.4&1.0&0.4&0.3&0.9&0.3&1.0&1.5&0.8&0.4&1.0&0.1&1.0&2.0&0.6&1.4&2.5&0.7 \\
\textbf{\texttt{ml}} & 0.2&0.7&0.1&0.7&1.2&0.2&0.7&1.5&0.4&0.8&0.9&0.5&0.6&1.3&0.3&0.5&0.8&0.5&0.2&0.5&0.1&1.0&2.0&0.1&1.1&1.9&0.4&1.0&1.6&1.4&0.3&0.6&0.3&0.9&1.5&0.3&0.4&0.2&0.4 \\
\textbf{\texttt{mni}} & 0.1 & 0.1 & 0.0 & 0.0 & 0.1 & 0.0 & 0.2 & 0.3 & 0.1 & 0.3 & 0.4 & 0.2 & 0.5 & 0.6 & 0.2 & 0.0 & 0.1 & 0.0 & 0.6 & 0.2 & 0.5 & 0.3 & 0.5 & 0.1 & 1.2 & 1.5 & 0.4 & 0.1 & 0.3 & 0.0 & 0.1 & 0.4 & 0.0 & 0.5 & 0.8 & 0.0 & 0.8 & 1.0 & 0.3 \\
\textbf{\texttt{mr}}  & 0.2&0.4&0.1&0.6&1.4&0.3&0.6&1.1&0.3&1.1&1.4&1.2&1.0&1.9&0.6&0.9&2.1&0.6&0.7&1.3&0.6&1.2&1.5&1.6&1.0&1.3&0.3&1.5&2.1&0.7&0.6&0.4&0.4&0.6&1.4&0.1&1.6&3.2&1.2 \\
\textbf{\texttt{or}} &  0.2 & 0.4 & 0.2 & 0.2 & 0.3 & 0.1 & 0.6 & 1.2 & 0.4 & 1.1 & 1.6 & 0.2 & 0.7 & 1.2 & 0.6 & 0.2 & 0.3 & 0.2 & 0.4 & 1.0 & 0.6 & 0.7 & 0.9 & 0.2 & 1.5 & 1.6 & 0.9 & 0.8 & 0.8 & 1.0 & 0.6 & 0.5 & 0.3 & 0.2 & 0.4 & 0.3 & 0.2 & 0.5 & 0.4\\
\textbf{\texttt{pa}} & 0.3 & 0.4 & 0.3 & 0.2 & 0.6 & 0.0 & 0.9 & 1.3 & 0.3 & 3.1 & 4.3 & 2.1 & 0.8 & 0.9 & 0.6 & 0.1 & 0.2 & 0.3 & 0.5 & 0.8 & 0.4 & 0.8 & 0.8 & 0.5 & 0.6 & 0.7 & 0.2 & 0.6 & 0.6 & 0.7 & 0.2 & 0.2 & 0.3 & 0.3 & 0.5 & 0.1 & 2.7 & 3.8 & 2.0\\
\textbf{\texttt{ta}} & 0.2 & 0.5 & 0.1 & 0.2 & 0.2 & 0.0 & 1.1 & 1.4 & 0.8 & 1.6 & 2.2 & 0.6 & 0.2 & 0.2 & 0.1 & 0.4 & 0.7 & 0.6 & 0.9 & 0.8 & 0.7 & 0.4 & 0.4 & 0.3 & 0.4 & 0.8 & 0.1 & 0.5 & 1.0 & 0.3 & 1.0 & 1.6 & 1.3 & 0.4 & 0.8 & 0.1 & 1.0 & 1.4 & 0.9  \\
\textbf{\texttt{te}} & 0.1 & 0.2 & 0.0 & 0.0 & 0.1 & 0.0 & 0.1 & 0.5 & 0.1 & 0.3 & 0.5 & 0.2 & 0.4 & 0.7 & 0.1 & 0.1 & 0.4 & 0.0 & 0.2 & 0.5 & 0.1 & 0.2 & 0.7 & 0.0 & 0.2 & 0.5 & 0.0 & 0.6 & 0.8 & 0.3 & 0.6 & 0.8 & 0.2 & 0.2 & 0.4 & 0.6 & 1.0 & 1.6 & 0.4 \\
\textbf{\texttt{en}} & 0.2 & 0.5 & 0.1 & 0.1 & 0.9 & 0.0 & 0.8 & 1.1 & 0.3 & 1.4 & 1.8 & 0.9 & 0.2 & 0.7 & 0.2 & 0.5 & 0.6 & 0.1 & 0.3 & 0.9 & 0.2 & 0.6 & 0.9 & 0.3 & 0.7 & 0.7 & 0.4 & 1.1 & 1.3 & 0.9 & 1.0 & 2.1 & 0.3 & 0.2 & 0.3 & 0.1 & 1.1 & 1.2 & 1.3 \\
\bottomrule
\end{tabular}
\vspace{-1ex}
\caption{Multilingual benchmark with IndicBART: standard deviations.}
\label{tab:multilingual-indicbart-std}

\vspace{3ex}

\setlength{\tabcolsep}{1.0ex}
\centering\scriptsize
\begin{tabular}{rrrrrrrrrrrrrrrrrrrrrrrrrrrrrrr} 
 \toprule
& \multicolumn{3}{c}{\textbf{\texttt{bn}}} & \multicolumn{3}{c}{\textbf{\texttt{gu}}} & \multicolumn{3}{c}{\textbf{\texttt{hi}}} & \multicolumn{3}{c}{\textbf{\texttt{ml}}} & \multicolumn{3}{c}{\textbf{\texttt{mni}}} & \multicolumn{3}{c}{\textbf{\texttt{mr}}} & \multicolumn{3}{c}{\textbf{\texttt{ta}}} & \multicolumn{3}{c}{\textbf{\texttt{te}}} & \multicolumn{3}{c}{\textbf{\texttt{ur}}} & \multicolumn{3}{c}{\textbf{\texttt{en}}} \\
\cmidrule(lr){2-4}\cmidrule(lr){5-7}\cmidrule(lr){8-10}\cmidrule(lr){11-13}\cmidrule(lr){14-16}\cmidrule(lr){17-19}\cmidrule(lr){20-22}\cmidrule(lr){23-25}\cmidrule(lr){26-28}\cmidrule(lr){29-31}
 & \multicolumn{1}{c}{R2} & \multicolumn{1}{c}{RL} & \multicolumn{1}{c}{BL} & \multicolumn{1}{c}{R2} & \multicolumn{1}{c}{RL} & \multicolumn{1}{c}{BL} & \multicolumn{1}{c}{R2} & \multicolumn{1}{c}{RL} & \multicolumn{1}{c}{BL} & \multicolumn{1}{c}{R2} & \multicolumn{1}{c}{RL} & \multicolumn{1}{c}{BL} & \multicolumn{1}{c}{R2} & \multicolumn{1}{c}{RL} & \multicolumn{1}{c}{BL} & \multicolumn{1}{c}{R2} & \multicolumn{1}{c}{RL} & \multicolumn{1}{c}{BL} & \multicolumn{1}{c}{R2} & \multicolumn{1}{c}{RL} & \multicolumn{1}{c}{BL} & \multicolumn{1}{c}{R2} & \multicolumn{1}{c}{RL} & \multicolumn{1}{c}{BL} & \multicolumn{1}{c}{R2} & \multicolumn{1}{c}{RL} & \multicolumn{1}{c}{BL} & \multicolumn{1}{c}{R2} & \multicolumn{1}{c}{RL} & \multicolumn{1}{c}{BL} \\
\midrule
\textbf{\texttt{bn}} & 0.6 & 0.9 & 0.6 & 0.3 & 0.2 & 0.5 & 0.5 & 0.5 & 0.2 & 0.8 & 1.8 & 0.5 & 1.1 & 0.6 & 1.0 & 0.5 & 1.0 & 0.9 & 1.1 & 0.9 & 1.0 & 0.5 & 0.8 & 0.4 & 1.5 & 1.4 & 0.9 & 2.1 & 1.6 & 1.2\\
\textbf{\texttt{gu}} & 2.0 & 1.8 & 1.2 & 0.4 & 0.8 & 1.4 & 0.6 & 0.4 & 0.4 & 2.3 & 2.5 & 1.1 & 0.3 & 0.7 & 0.2 & 1.6 & 0.7 & 1.3 & 1.1 & 0.5 & 1.0 & 1.8 & 1.9 & 0.7 & 0.9 & 0.6 & 1.1 & 1.7 & 1.1 & 0.8\\
\textbf{\texttt{hi}} & 1.3 & 1.5 & 0.6 & 0.9 & 1.2 & 0.5 & 1.6 & 1.6 & 1.3 & 1.7 & 1.7 & 0.6 & 0.3 & 0.7 & 0.3 & 1.4 & 0.8 & 0.4 & 1.1 & 1.1 & 1.4 & 1.9 & 1.7 & 1.2 & 1.2 & 1.0 & 0.8 & 2.4 & 1.3 & 2.3\\
\textbf{\texttt{ml}} & 1.1 & 0.8 & 0.5 & 0.6 & 1.3 & 1.3 & 0.3 & 0.1 & 0.5 & 1.8 & 0.8 & 1.2 & 1.5 & 2.0 & 1.3 & 2.4 & 1.9 & 1.7 & 2.7 & 2.0 & 2.4 & 1.8 & 1.6 & 0.5 & 0.7 & 1.0 & 1.0 & 3.6 & 3.0 & 2.4 \\
\textbf{\texttt{mni}} & 0.5 & 0.6 & 0.4 & 0.3 & 0.5 & 1.1 & 0.7 & 0.6 & 0.5 & 2.1 & 2.3 & 2.1 & 0.7 & 0.4 & 0.3 & 1.5 & 1.4 & 1.5 & 1.3 & 1.1 & 1.2 & 0.8 & 0.8 & 0.8 & 1.1 & 1.0 & 1.1 & 2.9 & 2.2 & 2.6 \\
\textbf{\texttt{mr}} & 1.5 & 1.5 & 1.0 & 1.2 & 0.6 & 1.2 & 0.8 & 1.0 & 0.9 & 1.8 & 2.2 & 0.9 & 0.6 & 0.6 & 0.7 & 1.8 & 1.1 & 1.9 & 1.2 & 1.4 & 1.4 & 1.0 & 0.8 & 0.7 & 0.5 & 0.7 & 0.3 & 0.9 & 0.6 & 1.1 \\
\textbf{\texttt{ta}} & 1.6 & 1.5 & 0.9 & 1.5 & 1.1 & 2.5 & 1.1 & 0.8 & 0.4 & 1.3 & 1.5 & 0.6 & 0.6 & 0.4 & 1.0 & 1.3 & 0.5 & 0.4 & 1.0 & 0.8 & 1.8 & 0.3 & 0.4 & 0.3 & 1.0 & 1.1 & 0.7 & 2.1 & 1.7 & 0.7 \\
\textbf{\texttt{te}} & 1.6 & 1.5 & 0.7 & 0.7 & 0.2 & 1.1 & 1.0 & 0.7 & 0.9 & 0.9 & 1.6 & 1.2 & 0.6 & 0.6 & 0.9 & 1.4 & 1.7 & 1.5 & 1.8 & 1.5 & 1.4 & 0.6 & 0.7 & 0.3 & 0.3 & 0.8 & 1.0 & 2.1 & 1.2 & 0.4 \\
\textbf{\texttt{ur}} &  1.2 & 1.3 & 0.1 & 0.8 & 0.6 & 1.6 & 1.0 & 1.0 & 0.9 & 1.4 & 1.6 & 0.6 & 0.7 & 0.5 & 1.1 & 1.2 & 0.6 & 0.7 & 0.1 & 0.6 & 0.4 & 1.8 & 1.8 & 1.0 & 1.0 & 1.1 & 0.6 & 1.5 & 1.1 & 1.4\\
\textbf{\texttt{en}} & 1.9 & 1.7 & 1.1 & 1.4 & 0.7 & 1.5 & 0.8 & 1.4 & 1.0 & 1.2 & 1.4 & 0.7 & 0.2 & 0.5 & 0.3 & 1.2 & 1.1 & 1.0 & 1.0 & 1.0 & 1.7 & 1.1 & 0.7 & 0.1 & 0.7 & 0.4 & 0.4 & 2.3 & 1.3 & 2.5 \\
\bottomrule
\end{tabular}
\vspace{-1ex}
\caption{Multilingual benchmark with mBART: standard deviations.}
\label{tab:multilingual-mbart-std}
\end{sidewaystable}

\FloatBarrier
\section{Datasheet for PMIndiaSum}\label{appendix:data_sheet}

\subsection{Motivation}

\noindent\textbf{Q:} For what purpose was the dataset created? (Was there a specific task in mind? Was there a specific gap that needed to be filled? Please provide a description.)

\noindent\textbf{A:} The dataset was developed to facilitate research on monolingual, cross-lingual, and multilingual headline summarization, specifically for Indian languages. Given the nature of multilingualism in India, cross-lingual summarization can greatly ease information access, but there is currently little availability in such datasets that can be used for researching the topic. Refer to Section~\ref{sec:intro} and Section~\ref{sec:related_works} for more details.
\\
\\
\noindent\textbf{Q:} Who created the dataset (e.g., which team, research group) and on behalf of which entity (e.g., company, institution, organization)?

\noindent\textbf{A:} Researchers from IIIT Hyderabad and the University of Edinburgh created the dataset.
\\
\\
\noindent\textbf{Q:} Who funded the creation of the dataset?
\noindent\textbf{A:} Please refer to the author list for the people who created this dataset; the funding sources for authors are stated in the acknowledgement section.
\\
\\
\noindent\textbf{Q:} Any other comments?

\noindent\textbf{A:} No.

\subsection{Composition}
\noindent\textbf{Q:} What do the instances that comprise the dataset represent (e.g., documents, photos, people, countries)? (Are there multiple types of instances (e.g., movies, users, and ratings; people and interactions between them; nodes and edges)? Please provide a description.)

\noindent\textbf{A:} Each instance in the dataset contains a news article and corresponding headline either in the same or different languages. Refer to Table~\ref{fig:example1} for data samples.
\\
\\
\noindent\textbf{Q:} How many instances are there in total (of each type, if appropriate)?

\noindent\textbf{A:} The dataset consists of 76,680 monolingual and 620,336 cross-lingual document-headline pairs. Please refer to Table~\ref{tab:cls_counts} for individual language pair counts.
\\
\\
\noindent\textbf{Q:} Does the dataset contain all possible instances or is it a sample (not necessarily random) of instances from a larger set?

\noindent\textbf{A:} The dataset consists of all instances derived from the raw data we gathered and processed.
\\
\\
\noindent\textbf{Q:} Is any information missing from individual instances? If so, please provide a description, explaining why this information is missing (e.g., because it was unavailable). This does not include intentionally removed information, but might include, e.g., redacted text. 

\noindent\textbf{A:} No.
\\
\\
\noindent\textbf{Q:} Are relationships between individual instances made explicit (e.g., users' movie ratings, social network links)? If so, please describe how these relationships are made explicit.)

\noindent\textbf{A:} Yes. Data instances are mostly independent of each other. Data instances of document-headline pairs in different languages from the same article convey the same content in the original news articles.
\\
\\
\noindent\textbf{Q:} Are there recommended data splits (e.g., training, development/validation, testing)? If so, please provide a description of these splits, explaining the rationale behind them.

\noindent\textbf{A:} Yes. Refer to Section~\ref{sec:data_split} for an explanation. The split information is in the data file itself. We also provide a code zip which has the script to create data splits linked in Footnote~\ref{footnote:data-url}.
\\
\\
\noindent\textbf{Q:} Are there any errors, sources of noise, or redundancies in the dataset? 

\noindent\textbf{A:} Our expectation is that the articles are professionally written with no errors, and we take measures to ensure that there are no redundancies by removing duplicated data. However, it is possible that extraneous HTML elements may introduce errors that remain unfiltered from the raw crawled data. It is not feasible to manually inspect all data instances or automatically identify this type of noise.
\\
\\
\noindent\textbf{Q:} Is the dataset self-contained, or does it link to or otherwise rely on external resources (e.g., websites, tweets, other datasets)? (If it links to or relies on external resources, a) are there guarantees that they will exist, and remain constant, over time; b) are there official archival versions of the complete dataset (i.e., including the external resources as they existed at the time the dataset was created); c) are there any restrictions (e.g., licenses, fees) associated with any of the external resources that might apply to a dataset consumer? Please provide descriptions of all external resources and any restrictions associated with them, as well as links or other access points, as appropriate.)

\noindent\textbf{A:} The dataset is self-contained. The dataset can be downloaded, used, adapted, and re-distributed without restrictions.
\\
\\
\noindent\textbf{Q:} Does the dataset contain data that might be considered confidential (e.g., data that is protected by legal privilege or by doctor–patient confidentiality, data that includes the content of individuals' non-public communications)? If so, please provide a description.

\noindent\textbf{A:} No, as all articles in the dataset are publicly available from the PMIndia's website. 
\\
\\
\noindent\textbf{Q:} Does the dataset contain data that, if viewed directly, might be offensive, insulting, threatening, or might otherwise cause anxiety? If so, please describe why. 

\noindent\textbf{A:} Unlikely, but we cannot completely rule this out because the data contains news articles from a governmental website.
\\
\\
\noindent\textbf{Q:} Does the dataset relate to people? (If not, you may skip the remaining questions in this section.)

\noindent\textbf{A:} Yes, the majority of the articles are about the Prime Minister of India who is a public figure. The data also contains news about other real people. 
\\
\\
\noindent\textbf{Q:} Does the dataset identify any subpopulations (e.g., by age, gender)? If so, please describe how these subpopulations are identified and provide a description of their respective distributions within the dataset.

\noindent\textbf{A:} While the dataset does not explicitly identify any subpopulations based on factors such as age or gender, it is possible that such information may be mentioned in the news articles themselves, such as when referring to specific individuals. As news articles are likely to include details such as gender, age, occupation, etc., it is possible that these factors may be indirectly associated with the data instances in the dataset. However, we do not have any explicit information on the distributions of these subpopulations within the dataset.
\\
\\
\noindent\textbf{Q:} Is it possible to identify individuals (i.e., one or more natural persons), either directly or indirectly (i.e., in combination with other data) from the dataset? If so, please describe how.

\noindent\textbf{A:} Yes, there are individuals' names present in the data.
\\
\\
\noindent\textbf{Q:} Does the dataset contain data that might be considered sensitive in any way (e.g., data that reveals race or ethnic origins, sexual orientations, religious beliefs, political opinions or union memberships, or locations; financial or health data; biometric or genetic data; forms of government identification, such as social security numbers; criminal history)? If so, please provide a description.

\noindent\textbf{A:} While it is unlikely that the dataset contains sensitive information that is not already public, there is a possibility that certain news articles in the dataset may contain details that could be considered sensitive. For example, news articles may refer to individuals' personal information. However, as our dataset is derived from a public governmental news website, any sensitive information that may be present in the dataset is likely to have already been publicly disclosed.
\\
\\
\noindent\textbf{Q:} Any other comments?

\noindent\textbf{A:} No.

\subsection{Collection process}
\noindent\textbf{Q:} How was the data associated with each instance acquired? (Was the data directly observable (e.g., raw text, movie ratings), reported by subjects (e.g., survey responses), or indirectly inferred/derived from other data (e.g., part-of-speech tags, model-based guesses for age or language)? If data was reported by subjects or indirectly inferred/derived from
other data, was the data validated/verified? If so, please describe how.)

\noindent\textbf{A:} The data is crawled from the Prime Minister of India website followed by processing. It is observable on the website. The data is reported directly by the website.
\\
\\
\noindent\textbf{Q:} What mechanisms or procedures were used to collect the data (e.g., hardware apparatus or sensor, manual human curation, software program, software API)? (How were these mechanisms or procedures validated?)

\noindent\textbf{A:} The data was scraped from a website using a specifically-design open-source crawler and parser. Please refer to Section~\ref{sec:data_acquistion} for more details. We have manually inspected a few samples to verify the sample quality and parallelism.
\\
\\
\noindent\textbf{Q:} If the dataset is a sample from a larger set, what was the sampling strategy (e.g., deterministic, probabilistic with specific sampling probabilities)? 

\noindent\textbf{A:}  The dataset is not sampled from a larger corpus.
\\
\\
\noindent\textbf{Q:} Who was involved in the data collection process (e.g., students, crowd workers, contractors) and how were they compensated (e.g., how much were crowd workers paid)?

\noindent\textbf{A:} The data set is crawled and processed automatically. The scripts to crawl and process data are partly open-source, and partly written by the authors of this paper.
\\
\\
\noindent\textbf{Q:} Over what timeframe was the data collected? (Does this timeframe match the creation timeframe of the data associated with the instances (e.g., recent crawl of old news articles)? If not, please describe the timeframe in which the data associated with the instances was created.)

\noindent\textbf{A:} The data was crawled in early 2023. The crawled data span articles published between 2014 and early 2023.
\\
\\
\noindent\textbf{Q:} Were any ethical review processes conducted (e.g., by an institutional review board)? If so, please provide a description of these review processes, including the outcomes, as well as a link or other access point to any supporting documentation.

\noindent\textbf{A:} No.
\\
\\
\noindent\textbf{Q:} Did you collect the data from the individuals in question directly, or obtain it via third parties or other sources (e.g., websites)?

\noindent\textbf{A:} The dataset was obtained from a third-party website that publishes data or information of individuals.
\\
\\
\noindent\textbf{Q:} Were the individuals in question notified about the data collection? (If so, please describe (or show with screenshots or other information) how notice was provided, and provide a link or other access point to, or otherwise reproduce, the exact language of the notification itself.)

\noindent\textbf{A:} No. The news articles are public.
\\
\\
\noindent\textbf{Q:} Did the individuals in question consent to the collection and use of their data? (If so, please describe (or show with screenshots or other information) how consent was requested and provided, and provide a link or other access point to, or otherwise reproduce, the exact language to which the individuals consented.)

\noindent\textbf{A:} No. The news articles are public.
\\
\\
\noindent\textbf{Q:} If consent was obtained, were the consenting individuals provided with a mechanism to revoke their consent in the future or for certain uses? (If so, please provide a description, as well as a link or other access point to the mechanism (if appropriate).)

\noindent\textbf{A:} N/A.
\\
\\
\noindent\textbf{Q:} Has an analysis of the potential impact of the dataset and its use on data subjects (e.g., a data protection impact analysis) been conducted? (If so, please provide a description of this analysis, including the outcomes, as well as a link or other access point to any supporting documentation.)

\noindent\textbf{A:} No.
\\
\\
\noindent\textbf{Q:} Any other comments?

\noindent\textbf{A:} No.

\subsection{Preprocessing, cleaning, labeling}

\noindent\textbf{Q:} Was any preprocessing/cleaning/labeling of the data done (e.g., discretization or bucketing, tokenization, part-of-speech tagging, SIFT feature extraction, removal of instances, processing of missing values)? (If so, please provide a description. If not, you may skip the remainder of the questions in this section.)

\noindent\textbf{A:} Yes, detailed in Section~\ref{sec:data_processing}.
\\
\\
\noindent\textbf{Q:} Was the ``raw'' data saved in addition to the preprocessed/cleaned/labeled data (e.g., to support unanticipated future uses)? (If so, please provide a link or other access point to the "raw" data.)

\noindent\textbf{A:} The ``raw'' data is saved but not made public at the moment. We plan to do so shortly.
\\
\\
\noindent\textbf{Q:} Is the software used to preprocess/clean/label the instances available? (If so, please provide a link or other access point.)

\noindent\textbf{A:} We uploaded a code zip which has these scripts. 
\\
\\
\noindent\textbf{Q:} Any other comments?

\noindent\textbf{A:} No.

\subsection{Uses}

\noindent\textbf{Q:} Has the dataset been used for any tasks already? (If so, please provide a description.)

\noindent\textbf{A:} The dataset has been used by ourselves to fine-tune existing models to perform the headline summarization task. See Section~\ref{sec:benchamrk_experiments} for more details.
\\
\\
\noindent\textbf{Q:} Is there a repository that links to any or all papers or systems that use the dataset? (If so, please provide a link or other access point.)

\noindent\textbf{A:} No.
\\
\\
\noindent\textbf{Q:} What (other) tasks could the dataset be used for?

\noindent\textbf{A:} The dataset can be utilized for a wide range of NLP tasks concerning languages in India: language modelling, named entity recognition, linguistic analysis, etc.
\\
\\
\noindent\textbf{Q:} Is there anything about the composition of the dataset or the way it was collected and preprocessed/cleaned/labeled that might impact future uses? (For example, is there anything that a future user might need to know to avoid uses that could result in unfair treatment of individuals or groups (e.g., stereotyping, quality of service issues) or other undesirable harms (e.g., financial harms, legal risks) If so, please provide a description. Is there anything a future user could do to mitigate these undesirable harms?)

\noindent\textbf{A:} Yes, we applied strict language filtering which removed code-mixing instances, however, it could be a common and valid language phenomenon in languages of India.
\\
\\
\noindent\textbf{Q:} Are there tasks for which the dataset should not be used? (If so, please provide a description.)

\noindent\textbf{A:} Our dataset, like any large dataset, has the potential to be used in harmful ways. One example of such harm is the creation of models that generate hate speech or fake news using the data. Additionally, as previously mentioned, the dataset may contain sensitive information that should not be used for discriminatory or harmful purposes. It is important to carefully consider the intended use of the dataset and ensure that it aligns with ethical and legal standards.
\\
\\
\noindent\textbf{Q:} Any other comments?

\noindent\textbf{A:} No.

\subsection{Distribution}

\noindent\textbf{Q:} Will the dataset be distributed to third parties outside of the entity (e.g., company, institution, organization) on behalf of which the dataset was created? (If so, please provide a description.)

\noindent\textbf{A:} Yes, the data will be free to the public to download, use, modify and re-distribute.
\\
\\
\noindent\textbf{Q:} How will the dataset be distributed (e.g., tarball on website, API, GitHub)? (Does the dataset have a digital object identifier (DOI)?)

\noindent\textbf{A:} The dataset is currently hosted on Hugging Face as indicated in Footnote~\ref{footnote:data-url}, as well as on Google Drive as a zip file.
\\
\\
\noindent\textbf{Q:} When will the dataset be distributed?

\noindent\textbf{A:} The dataset is available now.
\\
\\
\noindent\textbf{Q:} Will the dataset be distributed under a copyright or other intellectual property (IP) license, and/or under applicable terms of use (ToU)? (If so, please describe this license and/or ToU, and provide a link or other access point to, or otherwise reproduce, any relevant licensing terms or ToU, as well as any fees associated with these restrictions.)

\noindent\textbf{A:} Yes, the dataset is distributed under the CC BY 4.0 license.
\\
\\
\noindent\textbf{Q:} Have any third parties imposed IP-based or other restrictions on the data associated with the instances? (If so, please describe these restrictions, and provide a link or other access point to, or otherwise reproduce, any relevant licensing terms, as well as any fees associated with these restrictions.).

\noindent\textbf{A:} The original website states that ``the material has to be reproduced accurately and not to be used in a derogatory manner or in a misleading context. Wherever the material is being published or issued to others, the source must be prominently acknowledged.'' More information is available at \href{https://www.pmindia.gov.in/en/website-policies/}{https://www.pmindia.gov.in/en/website-policies/}
\\
\\
\noindent\textbf{Q:} Do any export controls or other regulatory restrictions apply to the dataset or individual instances? (If so, please describe these restrictions, and provide a link or other access point to, or otherwise reproduce, any supporting documentation.)

\noindent\textbf{A:} Not known.
\\
\\
\noindent\textbf{Q:} Any other comments?

\noindent\textbf{A:} No.

\subsection{Maintenance}

\noindent\textbf{Q:} Who is supporting/hosting/maintaining the dataset?

\noindent\textbf{A:} Authors of this paper.
\\
\\
\noindent\textbf{Q:} How can the owner/curator/manager of the dataset be contacted (e.g., email address)?

\noindent\textbf{A:} Via email or issues in the Hugging Face or GitHub repositories as Footnote~\ref{footnote:data-url}.
\\
\\
\noindent\textbf{Q:} Is there an erratum? (If so, please provide a link or other access point.)

\noindent\textbf{A:} No.
\\
\\
\noindent\textbf{Q:} Will the dataset be updated (e.g., to correct labeling errors, add new instances, delete instances)? (If so, please describe how often, by whom, and how updates will be communicated to users (e.g., mailing list, GitHub)?)

\noindent\textbf{A:} Currently there is no plan to update the dataset.
\\
\\
\noindent\textbf{Q:} If the dataset relates to people, are there applicable limits on the retention of the data associated with the instances (e.g., were individuals in question told that their data would be retained for a fixed period of time and then deleted)? (If so, please describe these limits and explain how they will be enforced.)

\noindent\textbf{A:} No.
\\
\\
\noindent\textbf{Q:} Will older versions of the dataset continue to be supported/hosted/maintained? (If so, please describe how. If not, please describe how its obsolescence will be communicated to users.)

\noindent\textbf{A:} There is no older version of the dataset. Its obsolescence will be communicated to users at the venue where it is held.
\\
\\
\noindent\textbf{Q:} If others want to extend/augment/build on/contribute to the dataset, is there a mechanism for them to do so? (If so, please provide a description. Will these contributions be validated/verified? If so, please describe how. If not, why not? Is there a process for communicating/distributing these contributions to other users? If so, please provide a description.)

\noindent\textbf{A:} Yes, they can freely do so by any means as long as they abide by the license.
\\
\\
\noindent\textbf{Q:} Any other comments?

\noindent\textbf{A:} No.

\end{document}